\documentclass[10pt,twocolumn,twoside]{article}

\usepackage{graphicx}
\usepackage{mathptmx}      

\usepackage[utf8]{inputenc}
\usepackage[greek,english,british,USenglish]{babel}
\usepackage{float}
\usepackage{ifthen}
\usepackage{xspace}
\usepackage{authblk}
\usepackage{wrapfig}
\usepackage{xstring}
\usepackage{booktabs}
\usepackage{csquotes}
\usepackage{fancyhdr}
\usepackage{footnote}
\usepackage{geometry}
\usepackage{hyphenat}
\usepackage{multicol}
\usepackage{multirow}
\usepackage{tabularx}
\usepackage{microtype}
\usepackage{marginnote}
\usepackage{subcaption}
\usepackage[T1]{fontenc}
\usepackage[all]{nowidow}
\usepackage{tablefootnote}
\usepackage[inline]{enumitem}
\usepackage[usenames,dvipsnames]{xcolor}
\usepackage[sort&compress,square,comma,numbers]{natbib}
\usepackage{pgfplots}
\usepackage{varioref}
\usepackage{hyperref}
\usepackage{cleveref}
\hypersetup{
	hidelinks=true,
	colorlinks=true,
	linkcolor={green!50!black},
	citecolor={blue!50!black},
	urlcolor={blue!80!black}
}

\makeatletter \def\cl@chapter{\@elt {theorem}} \makeatother

\newcommand{\rpar}{)} 

\newcommand\citereq[1]{%
	\ifthenelse{\equal{#1}{}}{%
		\textcolor{red}{[Citation Required!]}%
	}{%
		\textcolor{RedOrange}{[Need More Citations!]}%
		\cite{#1}%
	}%
}
\makeatletter
\newcommand\footnoteref[1]{\protected@xdef\@thefnmark{\ref{#1}}\@footnotemark}
\@ifundefined{url}{\newcommand\url[1]{#1}}{}
\makeatother

\newcommand{\formatmn}[1]{%
		\fontfamily{ptm}\selectfont%
		\bfseries\scriptsize#1%
}

\newcommand\athome{RoboCup@Home\xspace}

\makeatletter
\newcommand{\acronym}[3]{\newcommand{#1}[1][]{%
	\ifcsname @@#3\endcsname%
		\IfStrEqCase{##1}{%
			{n}{#3\marginnote{\formatmn{#3}}}%
			{f}{#2~(#3)}%
			{fn}{#2 (#3\marginnote{\formatmn{#3}})}%
			{x}{#2}%
			{xn}{#2\marginnote{\formatmn{#3}}}%
		}[#3]%
	\else%
		\IfSubStr{##1}{n}%
			{#2~(#3\marginnote{\formatmn{#3}})}%
			{#2~(#3)}%
		\expandafter\gdef\csname @@#3\endcsname{}%
	\fi\xspace%
}}
\makeatother

\makeatletter
\renewenvironment{abstract}{%
	\small%
	\if@twocolumn%
		\textbf{\abstractname} --- %
	\else%
		\begin{center}%
			{\bfseries \Large\abstractname\vspace{\z@}}
		\end{center}%
		\quotation%
		\fi%
}{%
	\if@twocolumn\else\endquotation\fi%
}
\makeatother
\acronym{\wrs}{World Robot Summit}{WRS}
\acronym{\erl}{European Robotics League}{ERL}
\acronym{\tc}{Technical Committee}{TC}
\acronym{\oc}{Organizing Committee}{OC}
\acronym{\tdp}{Team Description Paper}{TDP}
\acronym{\tdps}{Team Description Papers}{TDPs}
\acronym{\mt}{Machine Translation}{MT}
\acronym{\hri}{Human-Robot Interaction}{HRI}
\acronym{\asr}{Automatic Speech Recognition}{ASR}
\acronym{\nlp}{Natural Language Processing}{NLP}
\acronym{\nlu}{Natural Language Understanding}{NLU}
\acronym{\ssl}{Sound Source Localization}{SSL}
\acronym{\spl}{Standard Platform Leagues}{SPL}
\acronym{\opl}{Open Platform League}{OPL}
\acronym{\dspl}{Domestic Standard Platform League}{DSPL}
\acronym{\sspl}{Social Standard Platform League}{SSPL}
\acronym{\gpsr}{\emph{General Purpose Service Robot}}{GPSR}
\acronym{\srad}{\emph{Speech Recognition and Audio Detection}}{SRAD}
\acronym{\egpsr}{\emph{Endurance General Purpose Service Robot}}{EGPSR}
\acronym{\eegpsr}{\emph{Enhanced Endurance General Purpose Service Robot}}{EEGPSR}

\acronym{\amcl}{Adaptive Monte Carlo Localization}{AMCL}
\acronym{\ann}{Artificial Neural Network}{ANN}
\acronym{\arnn}{Artificial Recursive Neural Network}{ARNN}
\acronym{\asp}{Answer Set Programming}{ASP}
\acronym{\brisk}{Binary Robust Invariant Scalable Keypoints}{BRISK}
\acronym{\brief}{Binary Robust Independent Elementary Features}{BRIEF}
\acronym{\clafu}{Classification Fusion}{CLAFU}
\acronym{\cnn}{Convolutional Neural Network}{CNN}
\acronym{\cuda}{Compute Unified Device Architecture}{CUDA}
\acronym{\dnn}{Deep Neural Network}{DNN}
\acronym{\dof}{Degrees of Freedom}{DoF}
\acronym{\dwa}{Dynamic Window Approach}{DWA}
\acronym{\fpfh}{Fast Point Feature Histograms}{FPFH}
\acronym{\hatp}{Hierarchical Agent-based Task Planner}{HATP}
\acronym{\hfsm}{Hierarchical Finite State Machine}{HFSM}
\acronym{\hog}{Histogram of Oriented Gradients}{HOG}
\acronym{\hrsi}{Human-Robot Spatial Interactions}{HRSI}
\acronym{\icp}{Iterative Closest Point}{ICP}
\acronym{\ism}{Implicit Shape Models}{ISM}
\acronym{\mrpt}{Mobile Robot Programming ToolKit}{MRPT}
\acronym{\ork}{Object Recognition Kitchen}{ORK}
\acronym{\pcl}{Point Cloud Library}{PCL}
\acronym{\ransac}{Random Sample Consensus}{RANSAC}
\acronym{\sift}{Scale-Invariant Feature Transform}{SIFT}
\acronym{\slam}{Simultaneous Localization and Mapping}{SLAM}
\acronym{\surf}{Speeded Up Robust Features}{SURF}
\acronym{\svm}{Support Vector Machine}{SVM}
\acronym{\tld}{Track-Learning Detection}{TLD}
\acronym{\ukf}{Unscented Kalman Filter}{UKF}
\acronym{\yolo}{You Only Look Once}{YOLO}
\acronym{\laser}{Laser Range Finder Scanner}{LRFS}

\title{%
	\vspace{-2cm}%
	Trends, Challenges and Adopted Strategies in \athome%
	\\[-0.2cm]
	{\normalsize --- Extended version ---}
}
\author{Mauricio \textsc{Matamoros}\thanks{mauricio[at]uni-koblenz.de}}
\author{Viktor \textsc{Seib}\thanks{vseib[at]uni-koblenz.de}}
\author{Dietrich \textsc{Paulus}\thanks{paulus[at]uni-koblenz.de}}
\affil{\small Active Vision Group --- University of Koblenz}
\date{May 30, 2018}

\begin{document}
\maketitle

%
%

%
%
%

\begin{abstract}
Scientific competitions are crucial in the field of service robotics.
They foster knowledge exchange and allow teams to test their research in unstandardized scenarios and compare result.
Such is the case of \athome.
However, keeping track of all the technologies and solution approaches used by teams to solve the tests can be a challenge in itself.
Moreover, after eleven years of competitions, it's easy to delve too much into the field, losing perspective and forgetting about the user's needs and long term goals.

In this paper, we aim to tackle this problems by presenting a summary of the trending solutions and approaches used in \athome, and discussing the attained achievements and challenges to overcome in relation with the progress required to fulfill the long-term goal of the league.
Hence, considering the current capabilities of the robots and their limitations, we propose a set of milestones to address in upcoming competitions.

With this work we lay the foundations towards the creation of roadmaps that can help to direct efforts in testing and benchmarking in robotics competitions.

{%
	\vspace{0.5\baselineskip}
	\noindent%
	\textbf{Remark:} Information from 2018 participants couldn't be considered since the list of attending teams to Montreal 2018 and their scoring in the \athome leage was not available at the writing time of this manuscript.
}
\end{abstract}

%
%
%
\section{Introduction}
\label{sec:introduction}
In the eleven years since its foundation in 2006, the \athome league has played an important role fostering knowledge exchange and research in service robotics.
Moreover, nowadays the competition can influence --and sometimes direct-- the course of research in the area of domestic service robotics.

Having such impact is not a minor thing.
In consequence, the \athome league has the responsibility of planning carefully what needs to be tested and when to introduce changes by establishing milestones for the competition.
This, of course, can only be done after analyzing the grounds in which the league is standing.

In response, in this paper we present two main contributions.
First, we present a survey of the approaches and technical solutions reported by teams in each of the different basic functionalities or abilities to accomplish a task.
The conducted overview serves as a basis for our second contribution since it reveals to us the capabilities of the competing robots.
To achieve this, we tap on several information sources, including:
\begin{enumerate*}[label=\alph*)]
	\item claims made in the \tdps,
	\item relevant publications,
	\item rulebooks,
	\item multimedia material available on-line,
	and
	\item our cumulative experience as participants and referees in RoboCup@Home since 2009.
\end{enumerate*}

Second, we discuss the challenges yet to be overcome that we have identified throughout these first eleven years of \athome.
This discussion considers not only the robot's current capabilities, but the feats attained since the foundation of the league.
In addition, we also consider information from conducted polls targeting potential customers necessities all over the globe but specifically from people living in Germany, Japan, Mexico, and The Netherlands.
Consequently, we propose several sets of features, tasks, or applications that have to be addressed to achieve the goal of \athome, but making special emphasis in those relevant in the short term.
Nonetheless, it is important to point out that the discussion focuses exclusively on service robots (referred hereinafter as robots for simplicity), with particular application for domestic environments.
In the same sense, we are focusing only on \athome without ignoring the existence of related robotics competitions like the \erl and \wrs.

%
%
%

This manuscript is organized as follows:
In
\Cref{sec:athome}, we present a brief introduction to \athome and its history\xspace
(experienced competitors might want to skip this section)%
.
In
\Cref{sec:hardware}, we provide a brief summary of adopted hardware solutions.
Once studied physical constrains,
in
\Cref{sec:strategies}, we address the strategies and software solutions used to cope with the trials of the competition.
Later on, we discuss the extrinsic and intrinsic challenges to overcome which are presented in \Cref{sec:challenges-extrinsic} and \Cref{sec:challenges-intrinsic}, respectively.
Finally, in
\Cref{sec:conclusions} we close by presenting our conclusions.

%
%
%
\section{\athome}
\label{sec:athome}
The \athome league was created in 2006 with the goal of developing robots capable of realizing all domestic chores and bringing them from the labs into people's homes.
As stated in its website\footnotemark,
\enquote{the \athome league aims to develop service and assistive robot technology with high relevance for future personal domestic applications. It is the largest international annual competition for autonomous service robots and is part of the RoboCup initiative}.
\footnotetext{\label{footnote:robocup-website}Source: \url{http://www.robocupathome.org/} Retrieved: Jan 1st, 2018.}

\subsection{Organization}
\label{sec:athome-organization}
The competition takes place in a test arena.
This arena is configured to look like a typical apartment of the hosting country.
However, the specifications and appearance of the arena are kept undisclosed to participants until they arrive.

The competition normally lasts 6 days.
During the first day and a half, period known as setup, teams prepare their robots and familiarize them with the arena; while the last day is reserved for the final demonstration and the award ceremony.

In the competition, which is divided in two stages, each robot has to solve a set of tests.
In most tests, the robots have to solve a household-related task while their abilities and performance are evaluated.
Some tests may take place outside the competition arena.
However, the competition scenarios are never specified beforehand.
Finally, the best 50\% of the participants in Stage~I advance to the Stage~II, from which only the very best would advance to the Finals.

\subsection{Brief History of \athome}
\label{sec:athome-history}

During the first two years, the tests were scored with boolean criteria.
A team would receive points only if the robot successfully accomplished the given task.
This had a lot of setbacks, but the most important one was that it was barely impossible to analyze the robot's performance in each ability from the scores.
Therefore, a new scoring system was introduced in 2008 that is formally introduced and analyzed in~\cite{Wisspeintner2009}.
From this year on, tests were split in a sequential set of goals.
In this schema, a robot can't advance to the next step unless (successfully) completing the current one.
This allowed to delimit the degree of uncertainty, making possible to estimate performance by measuring the contribution of each ability to each reached goal; as performed by \citeauthor{Iocchi2015} in their analysis in~\cite{Iocchi2015}.
In addition, this schema considers increasing the difficulty every third year, fine-tuning only in between.
This provides teams with enough time to tune their algorithms for the newly introduced challenges.

\begin{figure*}[t]
	\centering
	\includegraphics[width=0.9\textwidth]{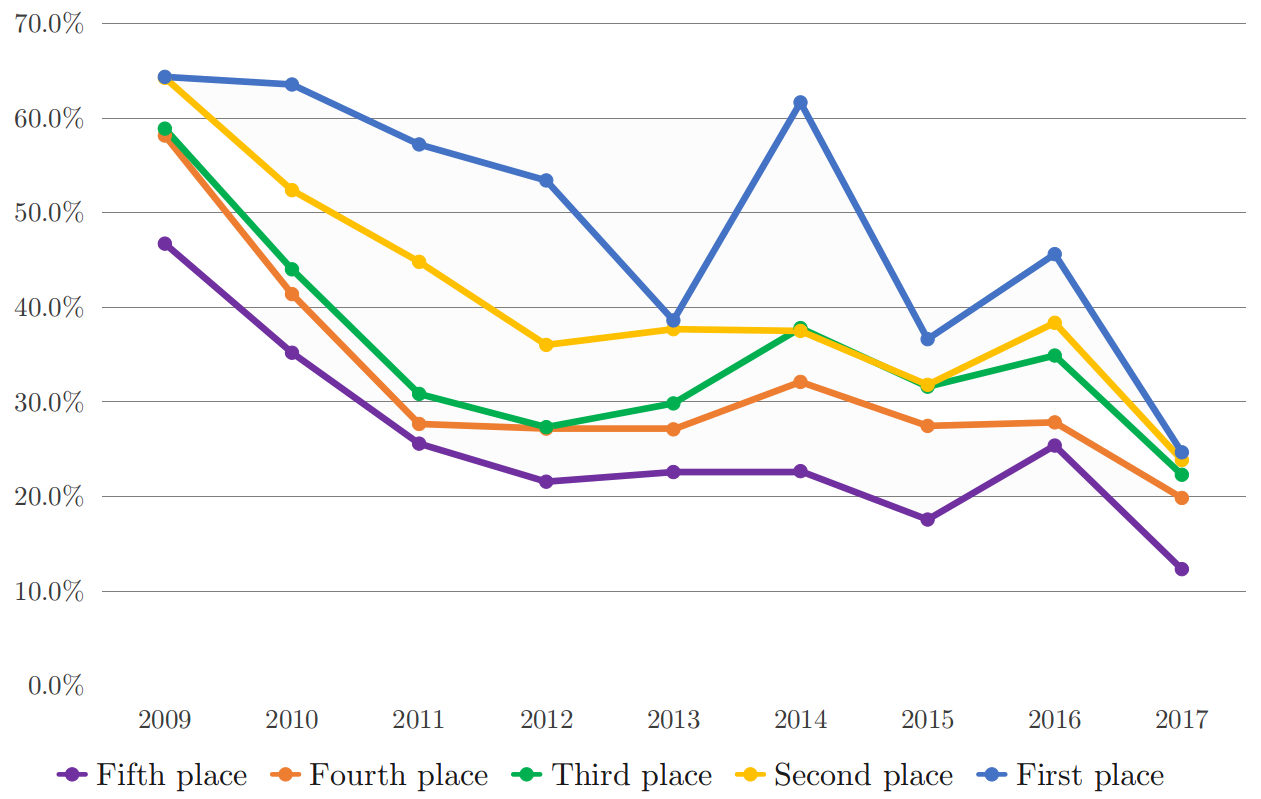}
	\caption{Performance of teams in the Top5 (2009 -- 2017)}
	\label{fig:yearly-performance}
	{\centering\footnotesize Obtained score with respect to the maximum attainable excluding special bonuses}
\end{figure*}

However, by 2013 the \tc had noticed a generalized decrease in performance (see~\Cref{fig:yearly-performance}).
Many robots failed in tasks that were considered more or less solved.
At the same time, it wasn't rare that robots couldn't even try\footnotemark.
For this reason, a test was specially designed to measure performance in basic skills for 2014.
Unfortunately, most of the robots did not come off well despite the relative simplicity of the benchmarking tasks.
Based on these results, the \tc decided to modify the test scheme proposed by~\citeauthor{Wisspeintner2009}~\cite{Wisspeintner2009} in Stage~I.

\footnotetext{Since 2010, the standard procedure involved waiting for the door to be opened, enter the arena, and retrieve a command.
Failing to detect when the door was opened, or misunderstanding the given command would make it impossible for the robot to advance any further.
Loose USB connectors and ambient noise promoted these problems even more.
This caused many teams with robust manipulation, people recognition, task planning, etc., to leave the arena without showing at what they excelled.}

Therefore, by 2015 tests with a sequential execution scheme were replaced with tests focused on measuring performance.
The former had the intention of attracting new participants by easing the Stage~I while the difficulty level of the Stage~II continued increasingft.
This new schema, in addition to include semi-isolated benchmarking of relevant abilities, also tried to tackle the luck factor by requiring each ability test to be run three times and considering only the average of the best two runs.
The tested abilities were
\begin{enumerate*}[label=\arabic*)]
	\item navigation, localization, and mapping,
	\item people recognition,
	\item people tracking,
	\item object recognition and manipulation,
	\item speech recognition and sound-source localization,
	and
	\item ability integration.
\end{enumerate*}

The introduced changes quickly paid off.
Besides providing benchmarking data to the \tc, the changes made evident to teams where their own weaknesses were.
This led to an immediate increase in performance in 2016 as~\Cref{fig:yearly-performance} shows.
Consequently, the \tc decided to raise once more the difficulty for 2017.
The gathered data was used to analyze strengths and weaknesses, and push in those areas that required more attention.
Notwithstanding, this new schema came also with several setbacks.
It widened the breach between stages, led to a performance decrease in tests relying on integration, and fostered score chasing.

Since 2015 a vast number of new challenges were introduced (many as bait) in the Stage~II with little success.
These new challenges include describing untrained objects, guiding people, avoiding tiny and light-reflecting objects, pouring, etc.
However, the newly introduced challenges were barely addressed.
Despite tests being sequential, the rules now allowed robots to receive help or skip a goal that wasn't a logic requirement to accomplish the next one.
As a consequence, participants went after easy to solve goals, completely ignoring the newly introduced challenges.
Increasing the reward didn't work either.
Nonetheless, \athome became more appealing to the audience since robots were allowed to show something, and team's frustration became almost inexistent.

Finally, 2017 introduced several important changes. \athome was split in three leagues:
\begin{enumerate*}[label=\arabic*)]
	\item the legacy \opl with no hardware restriction,
	\item the \dspl akin to the \opl, but using standardized hardware,
	and
	\item the \sspl that focuses in high-level \hri.
\end{enumerate*}
This division caused several veteran teams to migrate to the \spl{}, opening new spaces for new participants.
The former, in addition with the increase in the difficulty of several tasks led to the lowest performances in the history of \athome.

%
%
%
\section{Summary of Hardware Solutions}
\label{sec:hardware}

In this section we present a summary of the hardware configurations most used in the \opl.
We focus on five key aspects:
\begin{enumerate*}[label=\arabic*\rpar]
	\item RGB-D sensor model,
	\item drive mechanism of the base or locomotion type,
	\item number of \dof of the \textit{head}
	\item number of manipulators and their \dof,
	and
	\item number of \dof of the \textit{torso}.
\end{enumerate*}
The aspects have been chosen for their potential influence in the robot's performance, either by expanding its perception or interaction with the environment.

\subsection{RGB-D Sensor}
\label{sec:hardware-rgbd}
\begin{figure}
	\centering
	\includegraphics[width=\columnwidth]{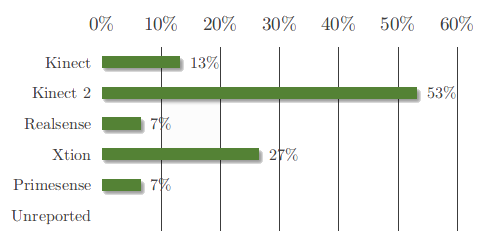}
	\caption{Most used RGB-D sensors (2017)}
	\label{fig:hardware-rgbd}
\end{figure}

As of 2017, all teams report the use of at least one RGB-D sensor.
From these, the preferred one seems to be the Microsoft Kinect 2 due to its incorporated Time of Flight sensor and better resolution, leaving the Asus Xtion in second place (see \Cref{fig:hardware-rgbd}).

\subsection{Base}
\label{sec:hardware-base}
To the date, all robots in \athome use wheels to move around.
Therefore, it can be considered that all robots have a base in which the driving mechanism lies.
Furthermore, there is no evidence in the \tdps pointing at a possible change of paradigm in the preferred locomotion type.

\begin{figure}
	\centering
	\includegraphics[width=\columnwidth]{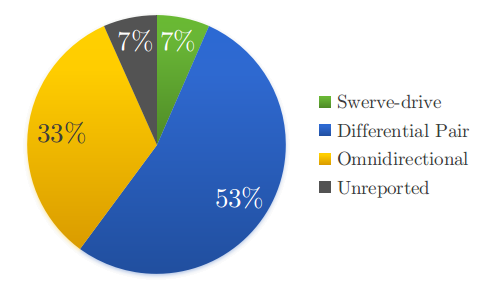}
	\caption{Reported locomotion types (2017)}
	\label{fig:hardware-base}
\end{figure}

From all possible configurations, the most used is the differential pair followed by omni-drive in either 3-wheeled or 4-wheeled configuration. There is only one reported use of \textit{Swerve drive}, a special type of omni-directional configuration in which all four wheels rotate independently. This information is summarized in \Cref{fig:hardware-base}.

\subsection{Head}
\label{sec:hardware-head}
\begin{figure}[t]
	\centering
	\begin{subfigure}[b]{0.48\columnwidth}
		\includegraphics[width=\textwidth]{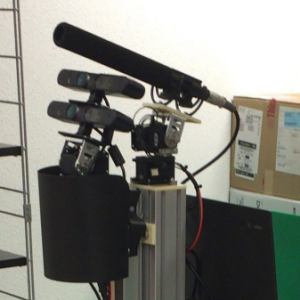}
		\caption{Exi@, Hibikino-Musashi}
		\label{fig:hardware-head-exia}
	\end{subfigure}
	~
	\begin{subfigure}[b]{0.48\columnwidth}
		\includegraphics[width=\textwidth]{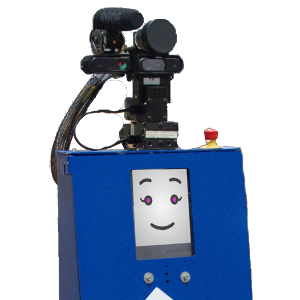}
		\caption{Lisa, homer@Uni-Koblenz}
		\label{fig:hardware-head-lisa}
	\end{subfigure}
	\caption{Addressing the right head may be deceiving.}
	\label{fig:hardware-head}
\end{figure}

To humans, identify the head of something in nature is normally easy, intuitive, and straightforward.
With robots, however, things change.
An unfamiliarized operator can get easily confused when speaking to a bicephalous robot (\Cref{fig:hardware-head-exia}), or a robot featuring a face located several centimeters away from its sensors (\Cref{fig:hardware-head-lisa}).

Hence, for the purpose of this summary we propose the following definition:

\begin{center}
	\begin{tabularx}{0.9\columnwidth}{ l X }
		\bfseries Robot head: & Unit comprising at least a camera and a microphone mounted on a pan-tilt unit. \\
	\end{tabularx}
\end{center}

Based on that definition, we are confident to declare that 67\% (10 out 15) of of the participant robots in Nagoya 2017 in the \opl featured a head.
The established standard in the league are heads with 2 DoF.

\subsection{Manipulator}
\label{sec:hardware-manipulator}
Few people conceive a domestic service robot with no arms.
In fact, most of the activities considered important by potential customers involve object handling.
Moreover, most homes are designed to optimize spaces, while still being accessible to their inhabitants.
Therefore, it would make sense that most robots had anthropomorphic configurations.

\begin{figure}[t]
	\centering
	\includegraphics[width=\columnwidth]{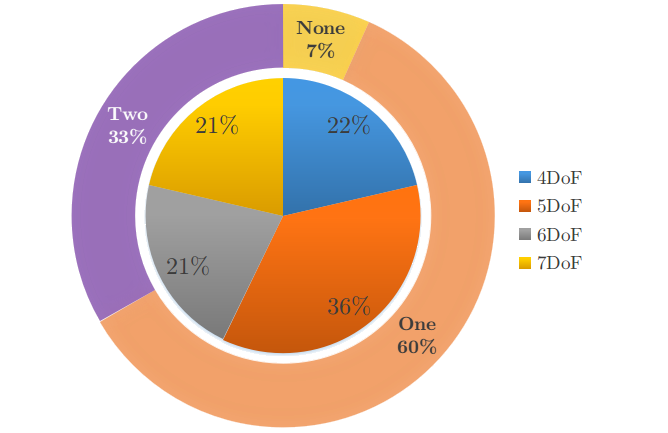}
	\caption{Adopted hardware solutions for manipulation}
	\label{fig:hardware-manipulator}
\end{figure}

However, to handle objects, most team use either home-made or proprietary low-cost hardware~\Citep{tobi2017,uchile2017}.
Although professional arms might be seen as the best option due to their strength and precision, their size makes them unfit for domestic narrow spaces.
Therefore, they are rarely used in \athome, with none of these present in Nagoya 2017.
In contrast, home-made manipulators are usually  anthropomorphic and much cheaper, although they lack the precision and strength of the former ones.
In Nagoya 2017 the number of \dof[x] for manipulators ranged from 4 to 7 with mode in 5 as \Cref{fig:hardware-manipulator} depicts.
Regarding the final effector strength, it is usually about $1.25kg$, insufficient to lift a $1.5L$ bottle of soda or cutting food.
Finally, since there is no mandatory task requiring two-handed manipulation, only few robots have more than one (see \Cref{fig:hardware-manipulator}).

\subsection{Torso}
\label{sec:hardware-torso}
Given an anthropomorphic configuration, a \textit{torso} would provide a robot with a panning and variable elevation for its head and upper limbs.
However, this is rarely the case.
In Nagoya 2017, only 9 out of 15 teams (60\%) reported a torso and, in all cases, it consisted of the elevator only.
However, in some cases, the elevator supported exclusively the manipulator.
Therefore, it is necessary to clarify that, in strict terms, units providing an additional \dof to the manipulator shouldn't be considered torsos.
Nonetheless, due to the lack of schematics, we decided to give our vote of confidence to teams, reporting the devices as alleged.

%
%
%
\section{Adopted Strategies and Software Solutions}
\label{sec:strategies}
In this section, we summarize the solutions most commonly adopted by teams to address each of the basic functionalities involved in the tests.

It is important to remark that none of the extremes in the abstraction layer is addressed.
On the high-level of abstraction side, all solutions differ, reflecting individual approaches of each team.
On the other side of the spectrum, low-level solutions are in most cases either unreported or often vendor-specific, without the influence or means to alter them by the teams.

%
%
%
\subsection{Frameworks and Middlewares}
\label{sec:strategies-frameworks}
Frameworks and middlewares can operate at many levels of abstraction.
This section refers exclusively to solutions for the intercommunication of the multiple modules that operate a service robot.
In addition, there is no further reference to the platforms imposed by vendors such as NaoQui.

Nowadays, ROS has become a tacit standard in robotics with all teams declaring its use in their \tdps for Nagoya 2017 and Montreal 2018~\Citep{Matamoros2018}.
Nonetheless, older frameworks like Orocos~\Citep{tue2017} are still in use.
Moreover, some teams still make good use of their own solutions~\Citep{pumas2017}.

That being said, it is important to remark that, although it has been discussed in the past, the \tc of the \athome league has discarded the idea of making ROS a compulsory standard.

%
%
%
\subsection{Audio, Speech, and Natural Language Processing}
\label{sec:strategies-speech}
The most broadly adopted solution to deal with speech consists in a pipeline.
In this pipeline,
a filtered audio signal feeds an \asr[f] engine to get a text-transcript for further processing.
Then, the transcript is sent to a natural language processor that extracts and conceals relevant information.
Finally, the acquired information is consumed by a high-level task planner that triggers the pertinent behaviors.
Although processing raw audio signals is technically possible \cite{Roy2002,Dominey2003}, this is still unexplored in \athome.

Typically, no external filters are used, leaving filtering to the microphone and the \asr engine~\cite{Doostdar2008}.
When reported, the most recurrent solution is HARK\footnotemark~\cite{aisltut2017,happymini2017,hibikinoOPL2017},
although it is intended for sound-source localization and separation.
\footnotetext{HARK (\textbf{H}onda Research Institute Japan \textbf{A}udition for \textbf{R}obots with \textbf{K}yoto University) is an open-source robot audition software that includes modules for \asr and sound-source localization and sound separation. Source: \url{https://www.hark.jp/}}

\begin{figure}
	\centering
	\includegraphics[width=\columnwidth]{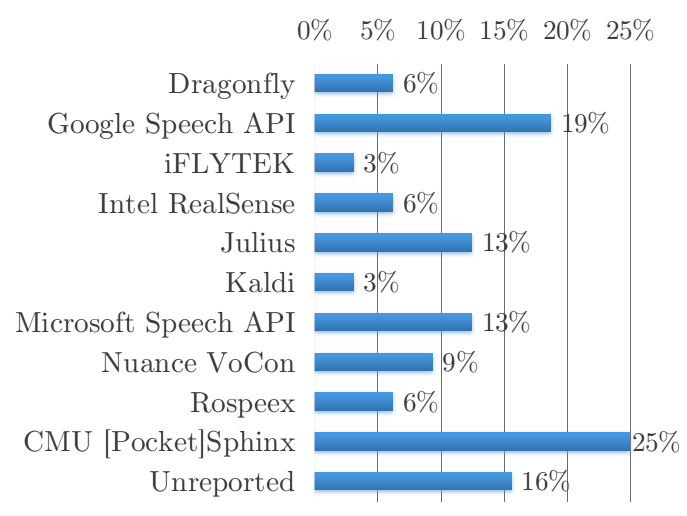}
	\caption{Trends in \asr[x] (2017)}
	\label{fig:asr-trends}
\end{figure}

Regarding \asr, the most commonly adopted off-line solutions include Julius~\cite{duckers2017}, the Microsoft Speech API~\cite{pumas2017}, and CMU Sphinx~\cite{kamerideropl2017,tobi2017} being the most popular solution as~\Cref{fig:asr-trends} shows.
Due to the limited computing power of their robots, most teams in the  \spl used cloud services, where Google speech API is the most popular approach~\cite{aisltut2017,jsk2017}.
However, network connectivity is often unreliable, a reason for which all teams using cloud services also have offline solutions as backup.

Moving forward to \nlp[x], despite the remarkable advances achieved in this area, little has been exploited in \athome.
To the date, we have found many teams still rely in keyword spotting and pattern matching to trigger the execution of a state machine~\cite{tinker2017,kamerideropl2017,Seib2015}, specially in simple tests.
In fact, \nlp and \nlu are mentioned in less than 50\% of the \tdps.
Nonetheless, robust A.I. solutions have always been in play and are now gaining strength.
Among the approaches for processing language, we found
probabilistic semantic parsers~\cite{utaustinvilla2017},
Multimodal Residual \dnn~\cite{wrighteagle2017},
ontology-based parsers over inference engines~\cite{pumas2017},
and
probabilistic parsers for syntax-tree extraction along with lambda calculus for semantic parsing~\cite{aupair2017}.
Of special mention are the Stanford Parser~\cite{stanford2011corenlponline}, the most broadly adopted solution for POS-tagging and syntactic tree extraction, and LU4R~\cite{Bastianelli2016}, a Spoken Language Understanding Chain for HRI developed in La Sapienza~\cite{spqrel2017} by participants of \athome which is also being used by several teams.
%
%
%
\subsection{Manipulation}
\label{sec:strategies-man}
How teams address manipulation heavily depends on the platform.
In the \sspl[x], due to limitations of the robot, manipulation is barely addressed and often skipped.
In direct opposition, the robot chosen for the \dspl[x] features a very precise and versatile manipulator that even incorporates a suction tip for lifting lightweight objects.
Last, but not least, manipulators in the \opl range from home-made to professional ones.

Implementation solutions for manipulation are often based on direct-inverse kinematic models with a closed-loop control and camera feedback as an alternative to the ROS manipulation stack.
However, nowadays many teams are migrating to \textit{MoveIt!}

Out of 32 participant teams in Nagoya 2017, manipulation was reported in only 18 (56\%) of the \tdps.
From these, 61\% are using \textit{MoveIt!}, while the rest relies on custom solutions that include
super-ellipsoid fitting~\Citep{tritonsdspl2017},
multiple deep auto-encoders fed with raw images and audio from the sensors~\Citep{erasers2017},
and
the built-in software solution provided~\Citep{spqrel2017}.

Other than picking and placing, robots are required to
open doors, pour liquids, scrub spots from a table, handle a tray, and move tableware.

In the \dspl, handling a tray is particularly problematic since the robot has only one manipulator, while opening doors doesn't seem to be a problem for the robot, and the suction tip comes handy for moving tableware.
In contrast, tray handling is bypassed in the \opl by mounting a custom tray in the robot, while the rest are usually skipped.
Unfortunately, no \tdp addresses the solution of any of these particular challenges.

Finally, it seems to be a growing trend in the use of deep-learning-based methods for manipulation, especially in planning.
In contrast with the traditional methods long-time used in industry, deep-learning-based approaches are computationally much more expensive, but also can be remarkably faster when a good-enough, non-optimal solution is acceptable.
In addition, their supporters claim they require much less expertise and effort to code.
Nonetheless, it is too early to know if these new methods will be a better solution for service robots.

%
%
%
\subsection{Navigation}
\label{sec:strategies-nav}
The basic functionality referred to as navigation in \athome involves four research areas, namely
\begin{enumerate*}[label=\arabic*\rpar]
	\item path planning,
	\item obstacle avoidance,
	\item localization,
	and
	\item mapping.
\end{enumerate*}
Navigation is fundamental in the competition.
It is assumed that all competing robots can safely navigate inside the arena.
Therefore, path planning, and localization in known environments (i.e. inside the arena) are considered to be solved, while obstacle avoidance is not.
Finally, on-line mapping becomes relevant only when robots are outside the arena, and nowadays is being extensively tested.

With a couple of exceptions, it can be said that all teams rely on the ROS navigation stack (see~\Cref{tbl:strategies-nav}).
However, it must be clarified that, in virtually all cases, this solution is adapted to the robot and the particular necessities of each team.
In this regard, the most broadly adopted solution sums up OpenSlam's Gmapping and \amcl with an $A^*$ path planner~\Citep{tue2017,uchile2017,tobi2017}.
Also a recurrent solution involves incorporating a Kalman filter to Gmapping~\Citep{tinker2017,walkingmachine2017}.
\Cref{tbl:strategies-nav} summarize these trends showing the reported ratio, and the most adopted solution with the percentage of reported use cases.

\begin{table}[t]
\caption{Trends in solutions for navigation. The first column, Report Ratio (R.R.) represents the percentage of \tdp in which the solution used was specified. The second column shows the most adopted solution and its use percentage with respect to the number of reports.}
	\centering
	\begin{tabularx}{\columnwidth}{ X c l }
	\toprule
	& \bfseries R.R. & \bfseries Mode (cases)
	\\ \midrule
	Path planning      & 63\% & $A^*$          (40\%)\\
	Obstacle avoidance & 56\% & ROS            (40\%)\\
	                   &      & Occupancy grid (28\%)\\
	Localization       & 63\% & \amcl          (60\%)\\
	Mapping            & 78\% & GMapping       (72\%)\\
	\bottomrule
	\end{tabularx}

\label{tbl:strategies-nav}
\end{table}

\subsubsection*{Path Planning and Obstacle Avoidance}
Other than $A^*$, several teams just report the use of ROS. In these cases, as well as when nothing is said, we assume teams use ROS' default behavior.
Other employed solutions include
randomized path planners~\Citep{aisltut2017},
and
wave-propagation algorithms based on the Fast Marching Method~\Citep{northeastern2017}.
Regarding obstacle avoidance, the most reported strategy is the use of occupancy grids.
To build the occupancy map, several teams take information from both, the laser range finder and the RGB-D camera.

\subsubsection*{Localization and Mapping}
These two abilities are often reported together.
Regarding localization alone, there are two reported solution differing from the robot's built-in localization and \amcl. These solutions are addressed by different teams and implement \slam using the \icp algorithm~\Citep{airobots2017,unsw2017}.
Both methods aim for accuracy and speed with limited resources.
On the mapping side, reported solutions other than GMapping include
hector \slam\Citep{duckers2017}, \mrpt and \icp\Citep{oittrial2017}, and Omnimapper\Citep{tritonsdspl2017}.

%
%
%
\subsection{Object Detection and Recognition}
\label{sec:strategies-or}
Object detection and object recognition are closely related, although they are often used separately.
Object detection is much faster than object recognition, so a common strategy while looking for objects involves performing a continuous detection until a potential area of interest is found.
A recurrent approach consists in taking the point-cloud of the RGB-D sensor and remove background, floor, and other surfaces (e.g. using Vectorial Quantization~\Citep{pumas2017} or \ransac~\Citep{northeastern2017}).
Later on, multiple color-depth images are extracted from the original RGB-D cloud using the detected clusters that can be further analyzed by the object recognizer.
This approach is also a computationally inexpensive alternative to deep-learning-based approaches that consume raw data from the sensors.

Most teams rely in more than one software solution to recognize objects, either running them in parallel and using some consensus algorithm, or implementing a process pipeline.
In addition, depth information is used for recognition based in contour and shape~\Citep{pumas2017,tritonsdspl2017,wrighteagle2017}.
However, there is no clear tendency on how the related modules are coupled.
For instance, some proposed solutions include
\yolo + \sift + \brief~\Citep{aisltut2017},
\surf + Continuous Hough-space voting + \ism~\Citep{homer2017},
color, size and shape histograms + SIFT~\Citep{pumas2017},
\pcl + \ransac + \yolo~\Citep{northeastern2017},
and
contours using LINEMOD + HSV color histograms + \surf considering joint models for occlusion~\Citep{wrighteagle2017}.

Detections and recognitions can take place either continuously or on demand.
For simplicity, and considering the default implementations of most used packages, it is more likely that both processes are running all the time.
However, unless the A.I. is designed to keep updating the world-model or knowledge-base of the robot with every new stimulus, acquired information will be simply discarded by the task planner.
In this regard, very few teams have reported the mechanisms to take advantage of continuous detection and recognition.

\begin{table}[t]
\caption{\small Adopted software solutions for object recognition. Presented results are based on the 81\% of reported strategies.}
\label{tbl:strategies-or}
	\centering
	\begin{tabularx}{0.8\columnwidth}{ X c }
	\toprule
	\bfseries Solution & \bfseries Reported uses
	\\ \midrule
	\yolo        & 23\%\\
	\sift        & 19\%\\
	\surf        & 15\%\\
	OpenCV Caffe & 12\%\\
	Tensorflow   & 12\%\\ \bottomrule
	\end{tabularx}
\end{table}

It is important to point out, however, that object recognition in \athome is not only about correctly labeling instances of previously trained objects.
Recognizing an object shape and orientation is, in many cases, fundamental for grasping, let alone stacking and storing.
Equally important are color, size, and relative position since
one can always refer to an object by description.
Therefore, robots must be able to identify features of untrained objects and correctly categorize them by likeliness.
Furthermore, since 2015 robots can be requested to describe objects they have never seen before.
Unfortunately, the strategies used to address the aforementioned challenges are not documented in the \tdps.

%
%
%
\subsection{People Detection \& Recognition}
\label{sec:strategies-pr}
People detection, recognition, and tracking are closely related but, at the same time, their approaches differ broadly.
In \athome, people detection and recognition means
localizing a relatively static target, while tracking people
involves a moving target (see~\Cref{sec:strategies-pt}).
This section relates only to detection and recognition
of people using exclusively visual information.

Regarding people detection, combining face and skeleton detection is the most popular approach since it reduces false positives and considers only people within range.
In addition, hybrid techniques like combining 3D object recognition with face detection (e.g. OpenFace), or analysis of thermal images~\Citep{Iocchi2015,uchile2015,tue2017} are also being used.
Focusing exclusively in face detection, all reported strategies are also recognition-capable, from which the more popular are
Openface~\Citep{tue2017},
Viola-Jones algorithm~\Citep{golem2017},
and
Haar-based algorithms~\Citep{happymini2017}.

\begin{table}[t]
\caption{Adopted software solutions for face detection and recognition. Presented results are based on the 67\% of available \tdps.}
\label{tbl:strategies-or}
	\centering
	\begin{tabularx}{0.8\columnwidth}{ X c }
	\toprule
	\bfseries Solution & \bfseries Reported uses
	\\ \midrule
	Openface              & 23\%\\
	Haar-based algorithms & 14\%\\
	Viola-Jones algorithm & 14\%\\
	Caffe                 & ~9\%\\
	Microsoft Face        & ~9\%\\
	OpenCV                & ~9\%\\ \bottomrule
	\end{tabularx}
\end{table}

On the other hand, all tests requiring to recognize a specific person either provide a description (see below), or provide the means to the robot to memorize the person's features and name.

Typically, getting to know a person for later recognition involves a facial recognition.
The solutions used by teams to achieve this are as diverse as the teams themselves.
Yet, some worth mentioning include
\hog descriptors with \svm classifiers~\Citep{alle2017},
strands perception people~\Citep{tobi2017},
Viola-Jones and eigenfaces~\Citep{golem2017},
Siamese \cnn{}s~\Citep{aupair2017},
and
Haar Cascades with either EigenFaces~\Citep{pumas2017} or \dnn~\Citep{happymini2017}.

Nonetheless, sometimes texture and color segmentation are also used as backup information.
It is, nevertheless, important to mention that cloud services are gaining popularity because of their robustness and ease of implementation.
However, they are also often unreliable due to connectivity problems so many teams prefer their own offline solutions.

When a description is provided, it may include height, gender\footnote{To avoid gambling, incorrect labeling is penalized, for which acceptable solutions are male, female and unknown.}, age\footnote{For simplicity, age estimation in \athome is fuzzy and maps to only three categories, child, adult, and elder.}, pose, relative position, and clothing.
To overcome these challenges, most teams rely on \dnn{}-based libraries and cloud services~\Citep{hibikinodspl2017,homer2017}.
Finally, it should be pointed out that some tasks may consist in requesting the robot to provide an accurate description of a person at the specified location.

%
%
%
\subsection{People Tracking}
\label{sec:strategies-pt}

People tracking is directly linked to two abilities: following and guiding people.
This sets another fundamental difference with object recognition: the robot can't see the face of the people being tracked.
Nonetheless, people tracking is always preceded by detection.
The robot has to memorize the person to be followed or guided, and indicate when that person can start moving.

For Nagoya 2017, many teams relied on the same approaches that were implemented back in 2006, like leg detection using the robot's \laser, and color segmentation with a probabilistic tracker~\Citep{Matamoros2018,Hans2008}.
Today, with better sensors and more powerful computers available, these methods have been improved and combined with other techniques.
Examples of these approaches include
leg-like clusterization with \svm and \ukf~\Citep{aisltut2017},
2D/3D leg detector fusing ellipse-fitting and 3D-windows obtained using a \dnn~\Citep{tritonsdspl2017},
and
\laser-based leg-tracker with an upper-body detector~\Citep{walkingmachine2017}.

Other approaches take advantage of the skeleton detection offered by RGB-D cameras~\Citep{unsw2017}.
However, this latter approach often produces unsatisfactory results due to vibrating sensors and people not fitting in the sensor's FoV.
On the other hand, interesting approaches involve
contour matching~\Citep{tue2017},
head tracking using a monocular camera and a Single Shot MultiBox Detector \cnn~\Citep{rtlions2017},
and
the \tld algorithm that learns from the difference between the estimated position of the target and the position detected in the analyzed frame~\Citep{airobots2017,tinker2017}.
Unfortunately, people tracking is one of the least reported functionalities in \tdps, with a frequency of 34\%.

Finally, at this point in time, there is no important difference between following and guiding.
For guiding, teams resort on the same techniques used for following, but using information from sensors looking backwards.

%
%
%
\section{Extrinsic Challenges}
\label{sec:challenges-extrinsic}
The challenges described in this section are considered to be extrinsic to the league.
Although they operate from outside, the influence they exert is big enough to be taken into account.
The discussed challenges are:
\begin{enumerate*}[label=\arabic*)]
	\item how to gain the general public interest and trust (\Cref{sec:challenges-people}),
	and
	\item how to manage available resources and safety constraints (\Cref{sec:challenges-resources}).
\end{enumerate*}

%
%
%

\subsection{Gaining People's Interest and Trust}
\label{sec:challenges-people}
Acquiring and maintaining the sympathy and interest of the general public is critic for research in service robotics.
Without people's trust, developing service robots would be pointless since there wouldn't be a market for them.
Furthermore, the active participation of people is fundamental in several related areas.

\subsubsection*{Engaging People}
\label{sec:challenges-people-engage}
It seems impossible to advance in natural \hri and \nlu without eventually taking advantage of people's innate skills.
For instance, latest advances in \asr and \mt were possible thanks to the huge amount of data for training that became available with the extended use of smart-phones.
However, the data acquired might be useless in \hri because it is very unlikely that people interact with robots in the same way they interact with their smart-phones.
Although there may be similarities, the difference in interaction becomes critical when it comes to data sets for deep learning.
In addition, while in a smart-phone a faulty transcription or an inaccurate translation are slight inconveniences, a robot that fails to correctly understand the given instructions becomes useless.
Moreover, having a good understanding of the operator's intentions is crucial in tasks that heavily depend on \hri.
Therefore, it is clear that instructing a robot must require significantly less time and effort than carry out the task.

The above suggests the need to build a corpus of interactions and, in particular, of spoken commands as a first step.
Steps towards that direction have already been made \citep{Bastianelli2014} and the produced corpus is being used in the \erl, although not yet in \athome\footnotemark.
The creation of corpora of that nature is not a trivial task.
First, people with experience in robotics, \asr, or \nlp cannot be involved in the process.
We have found that the expertise of people with a strong background in any of these fields acts as a biasing factor during the interaction process.
The expertise of experts makes them prone to alter pitch, speed, pauses, etc., when addressing robots.
What is worse, is that the biasing occurs at a subconscious level, making it extremely difficult for a person to prevent such behavior.
Therefore, non-experts shall be involved, and preferably volunteers with no experience in robotics and related fields.
The only requirement, at least in the beginning, is that the operator should have a friendly attitude towards the robot to ease the interaction.
\footnotetext{The generation and use of a corpus for \hri in \athome has been proposed and is discussed in \citep{Matamoros2018rc}.}

From these insights, the first logical step would be to attract the audience's attention.
However, this is not as straightforward as it may seem.
Unlike soccer, domestic chores are not exciting from most people's perspective.
In addition, soccer playing robots benefit from the familiarity of the audience who knows what is happening.
That's something we haven't achieved yet in \athome.
Most domestic chores are much more complex than any sport from a robotic's perspective due to the broadness of involved application domains of a domestic environment.
In consequence, robots perform their tasks slowly and fail more often, making the league even less attractive.
Therefore, new strategies are needed to attract audience's attention to \athome, getting them actively involved to produce the aforementioned corpora.

\subsubsection*{Regaining Trust}
\label{sec:challenges-people-trust}
\athome is directly affected by the \textit{Frankenstein Complex}.
While soccer and rescue robots are designed to operate in specific environments and circumstances, it is expected that service robots blend with people and use the same appliances.
This will be possible only if people consider them safe, even when using potentially harmful tools such as hammers, scissors, knives, and any other utensil of common use.

Media and entertainment industry are playing an important role in this matter.
For instance, most successful blockbusters and series related to robotics feature anthropomorphic machines that are hostile to humans.
Similarly, press and news constantly remind people that robots might not only turn evil, but also remark the high number of jobs that would be in jeopardy once robots are ready to be deployed.
Nonetheless, quite often informative services also make good press to the robots.
Moreover, sometimes their optimism harms what they are trying to exalt, raising expectations beyond what robots can actually do today.
We have witnessed on countless occasions the disenchantment of people who came to the competition to watch the prodigious robot of the news failing to grasp a simple object.

In both cases, competitions like \athome set the perfect opportunity to vindicate robots as the useful and innocuous servants they are.
It is during competitions, when people can effectively learn that robotics is a still-growing research area and why their participation is important to foster robot's development.
Furthermore, is in this kind of events where transdisciplinary networks can be established.
For instance, getting involved lawyers, sociologists, and philosophers can lead to the aperture of forums and research lines in topics of interest such as
licensing and patents,
liability in accidents caused by intelligent robots,
role of robots as \enquote{affective artifacts},
mid- and long-term impact of the integration of robots in human societies,
when a machine can be considered conscious,
and
rights and freedom of conscious robots.
Besides presenting the potential aperture of new research line in their fields, such forums can also contribute to engaging general public with the \athome community, preparing the way for the future.
We believe, in addition, that it's on our best interest to start discussing such matters sooner than later.

\subsubsection*{Conclusion}
\label{sec:challenges-people-conclusion}
The League's organizers are aware of these challenges and are working to overcome them.
One proposed solution is to have an expert moderating tests and explaining what is going on and why it is difficult.
In our opinion, another possible solution could be to show to the audience the goal and challenges of the running test to make it more appealing and understandable.

To conclude, more interdisciplinary research is needed to find solutions to integrate robots into human society.
It is also necessary to find ways to promote people's collaboration with the league to foster research in \nlu and \hri.
Often, hard questions come out, such as why should one acquire a robot while human assistants are much more efficient, cheaper, and hiring them helps creating jobs.
Therefore, it is important to find ways to vindicate robots as the useful and innocuous servants they are, and not present them as a threat to humanity.
%
%
%
\subsection{Available Resources and Safety Constraints}
\label{sec:challenges-resources}
In robotics exists a close relationship between the available resources and a reliable safe execution.
Today, with the internet becoming available everywhere, it becomes hard to resist to connect robots to the cloud.
However, such decisions must be addressed with care since, unlike with other technologies, a faulty execution can lead to a human being harmed.

\subsubsection*{Available Resources}
As the difficulty of the tests increases, it can be expected that the computational power required to solve them grew proportionally, especially when introduced changes are small.
It is true that it is impossible to tell what \textit{proportionally} means, especially in computer sciences and particularly in robotics.
In this regard, even the slightest changes, neglectable from a human perspective, can increase the complexity of a task in several orders of magnitude for the robot.
Nonetheless, in recent years we have detected a disproportional increase in the amount of computation needed to solve barely modified tasks.
Moreover, we have found that solving similar tasks requires more resources today than it required half a decade ago.

We believe this disproportion is partially due to the popularization of \ann and deep-learning-based solutions.
Although it is true that these new approaches are often more robust and normally easier to deploy, it is also true that they require a huge amount of computational power.
Besides, the enormous leap they caused in some disciplines have not fully reached the competition yet.
Therefore, we are expecting to witness some remarkable advances in the open demonstrations\footnotemark in coming years.
\footnotetext{The open demonstrations differ from regular tests in that they do not follow a script nor is there a predefined task to complete. Instead, teams can show whatever they want.}

Notwithstanding, experimenting with these technologies requires a computational power that laptops can't offer.
In addition, robots must become what smartphones are today: affordable devices providing reliable long-term operation with very limited resources.
This latter premise makes inadvisable to have robots performing heavy calculations all the time.
In conclusion, robot's computers might be unfit for most \dnn.
In consequence, roboticists look at cloud services.

\subsubsection*{Safety Constraints}
Cloud services offer several advantages besides reducing the amount of process a robot has to perform and increase the duration of its battery.
They also process information remarkably faster while taking advantage of the data provided by their users to enhance.

This, however, comes with some risks.
First, robots might experience a performance decrease due to lag and bandwidth problems during Internet rush-hours.
This might make them slow and inefficient in the best case but, in worst-case scenarios, they might endanger people's lives while waiting for the server's answer.
This becomes critical when performing nursing or baby-sitting tasks, in which a delay of a couple of seconds can cause severe harm.
Moreover, during emergencies the situation worsen.
Having all robots becoming inoperative due to a disconnection caused by a power shortage could be just an annoyance.
However, if the power shortage occurs during an earthquake or a fire, the mere presence of robots could hinder the evacuation instead of provide assistance.

Finally, and also a matter of security, but in a different sense, robots permanently connected to the internet have the potential risk of opening a window through which a malicious user could see and hear.
Nonetheless, implications on this matter won't be analyzed here any further.

\subsubsection*{Conclusions}
We believe that the robot's primary function should be helping people to have a better life.
This necessarily implies that robots must never compromise people's security and integrity.
Furthermore, the presence of a robot should help people to feel safe.
In consequence, minimum performance standards must be established to ensure a reliable and safe operation.
\athome has already taken the first steps by allowing cloud services and external computers on the condition that robots must remain operational and have full obstacle avoidance.

%
%
%
\section{Intrinsic Challenges}
\label{sec:challenges-intrinsic}
The challenges described in this section are considered to be intrinsic to the league.
Most of them largely correspond to the role of the league in directing the research and development in robotics.
First, we start discussing each of the most relevant abilities:
navigation~(\Cref{sec:challenges-nav}), people recognition~(\Cref{sec:challenges-pr}), people tracking~(\Cref{sec:challenges-pt}), object recognition~(\Cref{sec:challenges-or}),
manipulation~(\Cref{sec:challenges-man}), and \nlu~(\Cref{sec:challenges-asr}).
Finally, we close by addressing the the necessity of planning tasks to be tested (\Cref{sec:challenges-roadmaps}).

%
%
%
\subsection{Navigation}
\label{sec:challenges-nav}
In general terms, navigation can be split into two categories: indoor and outdoor navigation.
Although closely related, the categories need to be approached with disjoint ability sets.
Therefore, they are discussed separately.
Likewise and as in~\Cref{sec:strategies-nav}, navigation also involves obstacle avoidance, localization, and mapping.

\subsubsection*{Indoor-Navigation}
\label{sec:challenge-nav-in}
Despite the continuous improvements, nowadays robots won't make it in most homes.
Steps, wet floors, rough carpets, and in general uneven surfaces are challenges to overcome that have not been addressed yet, let alone houses with staircases.
Another minor detail important to consumers is quietness.
We have found that people prefers robots that move silently and without damaging the floor, specially if they are meant to clean the house during the night.

Other than stairs, big houses and facilities present new challenges, specially when it comes to localization.
The perception range of most robots is typically limited to about 4 meters, a distance short enough to impede localization in wide spaces and long corridors.
Furthermore, the map of the environment is often a given condition (e.g. during startup) that uses a remarkable amount of geometrical data.
Without semantic localization, robots might have it hard to localize themselves in buildings such as apartment towers and hospitals.
Thus, without a pre-existing map, both, mapping and localization, can take advantage of robot's ability to read.
This might prove to be specially useful in the long-term should a room or person be temporarily relocated.

Other important abilities that have been addressed in the past but haven't been solved\footnotemark~include
using an elevator, navigate in narrow corridors, and move furniture around.
The latter is of special importance when cleaning and tidying up rooms.
A robot should be able to move unattended objects that are blocking its path.
\footnotetext{Here it is important to remark that the aforementioned abilities haven't been solved in \athome. That does not necessarily imply that they are not being extensively tested by research groups. For example, \citeauthor{utaustinvilla2017} claim in~\citep{utaustinvilla2017} that their robot can localize itself after using an elevator.}

Last but not least, is functional touching.
People often rely on their body to move or stop objects when moving around.
For instance, it is common to see a person, carrying a lot of thing without a free hand, pushing a door with their hips, or holding a door with their foot.
In \athome, we call this kind of interaction \textit{functional touching} to differentiate it from collisions and intended manipulations.
Although allowed in the rulebook, functional touching hasn't been addressed yet by any team.

\subsubsection*{Outdoor-Navigation}
\label{sec:challenge-nav-out}
Either to take out the garbage, or when going to buy groceries, eventually robots will be required to go outside.
Urban environments come with a whole new set of challenges to overcome other than dealing with the elements.
The first one affects directly the robot's ability to move.
If in interiors the presence of unevenness in floors presents challenges, the terrain diversity outdoors seems more adverse.
Conducted polls reveal that there is a potential market for robots in pet owners, specially when it comes to exercise the animals.
Implementing such feature would require robots to be capable of moving in rough terrain like grass, sand, or gravel.

However, not only the unevenness increases outside.
Distances, and the number of stimuli to process, increases by a huge amount.
To localize themselves in open spaces, robots would require not only the ability to see objects at 20 meters and beyond, but it would be also necessary to deal with occlusions, and correlate the perceived information in real time.
As with autonomous cars, localization using point-clouds would be unreliable but, at the same time, the size difference and low movement speed would make it extremely hard to rely exclusively on GPS data.
Therefore, we foresee many situations in which a robot will not only need to recognize signs, but also read street and shop names in order to reach its destination.
In addition, in crowded streets a robot could encounter hundreds of pedestrians, including children, elders, dogs, and other robots, which it would have to evade.

Furthermore, speed is another important factor to take into account.
While indoors, a robot can take its time to solve a task, outside, the world imposes its own schedule.
Streetlights, public transport, and automatic doors are three examples of elements that require a quick response.
Therefore, the reaction time or robots must be improved, even in conditions of insufficient information.

\subsubsection*{Conclusions}
\label{sec:challenge-nav-conclusions}
Although \athome currently features wheeled-robots only, it is evident that this design is not suitable for most human environments.
Therefore, we will have to choose between making our environments robot-friendly, or provide robots with better means of movement.
Otherwise, robots will be restricted to those areas in which they can safely operate.

On the other hand, people can immediately recognize semantic information in their surroundings, having an innate ability to correlate the current environment with those previously known.
Eventually, robots will need to integrate semantic mapping (or an equivalent ability) capable of mimicking such human abilities in order to efficiently integrate in human environments.
%
%
%
\subsection{People Detection \& Recognition}
\label{sec:challenges-pr}
Detecting and recognizing people is fundamental for \hri[x] (\hri), therefore, it should be extensively tested in \athome.
Moreover, science-fiction has raised people's expectations, depicting robots as entities with an infallible, endless memory.
In consequence, most users expect robots capable of remember or find anyone.

\subsubsection*{Advances and Current State}
Although people detection is being extensively tested since it is involved in all types of \hri, people recognition is not.
Robots need to find a person to retrieve a command, deliver an object, answer a question, and detect a gesture, to name some examples.
However, most of the time the detection occurs with the person standing.
As of 2017, finding people lying or sitting was still problematic for most competitors.

In contrast, recognizing people hasn't bee fully addressed.
For our analysis, this feature is split in
\textit{people recalling}, when the robot needs to recall a person's name or order,
and \textit{people identification}, when the robot needs to identify a person's specific features.

Recalling, as stated in~\Cref{sec:strategies-pr}, usually consists in pairing the person's face with a name and, in some cases, some additional information like an order.
This ability has been tested since the very first competitions, but always with a reduced number of people (less than five).
In addition, although there shouldn't be a problem keeping the information longer, the face-name pairing lasts no longer than the test.

On the other hand, people identification is relatively new.
Robots can be requested both, to describe a group of people, or find a person matching the given description (relevant features are described in \Cref{sec:strategies-pr}).
However, by 2017 the tests addressing this ability either randomized the feature selection, or left its selection to the robot.
From our perspective, while most robots succeeded in finding their target, provided descriptions weren't very accurate, leading to the assumption that apart from relying in identification, heuristics were used to maximize the robot's chance of success.
In consequence, it is impossible to know precise information regarding general performance.

Nonetheless, it must be acknowledged that most robots perform good at counting people grouped for a photo.

\subsubsection*{Next Steps}
Considering the important number of features introduced in recent years, we believe the competition should focus in extending their usage.
Moreover, features like estimated age, gender, and relative position can be used to test awareness, decision making and planning.
Notwithstanding, focusing exclusively on detection and recognition, like with objects, the detection range needs to be increased, as well as the number of people in the memory of the robot.

On the features to be introduced in later stages and of importance in \hri we consider:
emotions and moods, activities, health and vital signs (inebriation, fatigue, sickness, sleep, etc.), skin and hair color, clothing names and styles, and identification by voice.
Also important, and shared with object recognition is addressing occlusions and translucent surfaces.

\subsubsection*{Conclusion}
Much has been achieved in the past eleven years.
However, \hri hasn't been tested as extensively as intended.
Furthermore, several tests require to be redesigned to ensure benchmarking data is available for analysis.

Finally, we find that visual and auditory information are equally important when addressing people.
Therefore, the integration of audio features in people detection and recognition has to be carefully planned.

%
%
%
\subsection{People Tracking}
\label{sec:challenges-pt}
People tracking is one of the first functionalities tested in \athome.
Although not necessarily useful when carrying out chores, it becomes relevant in daily life social events (e.g.~parties) and during shopping.
Its importance grows in non-domestic environments such as museums, hospitals, and restaurants.

\subsubsection*{Advances and Current State}
People following has been greatly improved since 2006.
Every year robots require less training time and are less prone to loose their target.
In addition, they can robustly handle people crossing or standing in between.
Moreover, back in 2014 robots were able to follow a person to  an elevator, get inside, and continue following after leaving the elevator.
And not only that, should the target go through a compact crowd blocking the sight, robots were able to go around it and meet their target.
Whether robots are still capable of doing it is unclear, since elevators and crowds are not part of the tests since 2015.

Guiding, on the other hand, was first tested in 2006 with little success, leading to its removal by 2008.
Since its reintroduction in 2015, the league has achieved good performance.
Furthermore, this ability is tested along with navigation.
After reaching a previously unknown location, the robot has to go back, find someone, and guide them to the location it just learned.

\subsubsection*{Next Steps}
To humans, it is unnecessary to study someone in order to follow them.
Moreover, it takes us only a quick glance to get all information required to track or even chase an object, let alone people.
Therefore, one of the first improvements in this basic functionality would be removing training time, so it's important to start reducing it.

Another important aspect is speed, which must start to be gradually increased.
Nowadays, robots follow and guide only \textit{professional walkers}, who walk at constant pace to keep tests fair.
In addition, professional walkers are robot-friendly, instructed to follow the instructions of the robot and even slow down when it has lost track.
However, people usually walk a lot faster than professional walkers, and are more likely to leave the robot behind if it can not follow their pace.
Also, when following and guiding, people have no inconvenience in sliding through crowds or narrow spaces, temporary losing contact to each other.
In consequence, it is important that robots can keep track of people walking fast in crowded environments and be able to predict or estimate local rendezvous points after optimizing trajectories.

Also of relevance in people tracking is side-by-side walking.
When two people walk side by side, it is hard to tell who is guiding the march.
It is one of the most common activities performed by humans.
Thus, it is expected that robots learn how to walk with humans and not in front or behind them.

In direct relationship, walking holding hands is a natural way of interaction.
Moreover, it is of special importance when interacting with children.
Notwithstanding, physical contact is strictly forbidden in \athome for security reasons.
Therefore, this also needs to be addressed, but with caution.

\subsubsection*{Conclusion}
Although unrelated to most domestic chores, people following and guiding are necessary abilities for many other applications.
Yet, both abilities can be analyzed as a hybrid problem, namely people tracking and navigation.
Moreover, tracking can be made extensive to animals and any kind of visual objects.
Thus, tracking gets into the domain of object recognition, for which it becomes important to find the pertinent features that make it possible for robots to track something.
Nonetheless, other problems like reducing training time and increasing speed need to be addressed with a carefully designed strategy.

%
%
%
\subsection{Object Detection and Recognition}
\label{sec:challenges-or}
Mankind has transformed its environment giving priority to the sense of sight.
Therefore, in order to effectively blend in human society, robots must be able to decode the visual stimuli of the world we have built around us.
Furthermore, and from a totally utilitarian point of view, perceive objects is necessary for manipulation, a key feature to perform any chore.

\subsubsection*{Advances and Current State}
\label{sec:challenges-or-advances}
Despite the remarkable advances in computer vision in the recent years, anyone who has witnessed recent competitions could safely argue that robots are shortsighted.
On the one hand, is very unlikely that a robot can detect an object more than 4 meters away.
On the other hand, robots continuously fail to identify objects lying in direct line of sight.
Both situations would deem a person as visually impaired and unable to perform in most environments without the proper aids.
The former is mostly due to the short range and limited resolution of used RGB-D sensors.
Nonetheless, this allows robots to do much more than years ago when sensors were limited to VGA cameras and Time-of-Flight sensors with a resolution of only a few thousands of pixels.

Limitations aside, successful detections are also used to update the grid of an occupancy map.
This is an important step further from when visual information wasn't used for obstacle avoidance.
In the case of recognition, features from several sensors are extracted and stored.
Later on, the features of each detected candidate are analyzed (e.g.~as in color and shape detection) or evaluated against the features of all trained objects, depending on the task requirements.

However, objects in the arena are always placed sparsely, and with partial occlusions at most.
Also, the set of objects the robot should know is capped at 25, a very small quantity considering the number of objects people use to deal with at home.
Furthermore, although robots have to identify objects \textit{of a kind} (e.g.~an apple, regardless the color, size and texture of the fruit), such recognition is still on very early stages.

\subsubsection*{Next Steps}
\label{sec:challenges-or-next-steps}
Virtually all reported strategies process the input stream frame-by-frame, computing in a feed-forward manner.
In other words, to track an object, a robot has to identify it and calculate its position in each frame with the possible assistance of a filter.
Furthermore, the extracted visual information once used is discarded (i.e.~not used for reinforcement).
In contrast, advances in neurosciences suggest that the human brain uses contextual information to build scene representations, and it can choose to rely on memory search instead of visual search~\Citep{Oliva2004,Oliva2007}.
Having robots using context to perform object detection and recognition, is an interesting area to explore.
For instance, a robot could analyze a scene from a different perspective, calculating the positions of those objects that have been already identified and removing all their clusters from the frame, focusing only in those with a low recognition confidence.

Following the line of recognition as instance-class matching, there are more problems to address, like
increasing the number of objects to hundreds and beyond, perform recognition of stacked objects, discriminate two identical objects based on its relative position,
recognize translucent and transparent objects, and recognize objects in odd lightning conditions like direct sunlight and in the dark.

In a broader perspective, detection and recognition can be taken further on.
Identifying the orientation of an object prior to grasping, as well as the best location for placement, or even infer the weight of an objects are requirements yet to be addressed.
Moreover, quite often robots will need to deal with occlusions and objects behind crystals, identify dirt, dust and spots on the floor, and objects that change over time (e.g.~food).

\subsubsection*{Conclusions}
\label{sec:challenges-or-conslusions}
Object recognition applied to robotics is in a very early stage.
Similar to other abilities, a roadmap is necessary to direct advances on object recognition.
However, recognizing things, more than characterizing features, is an A.I.-complete problem.
The first logical steps would be those that help with manipulation, like identifying placing surfaces, grasping orientations, and containers.
Other useful features that can be addressed first include occlusions and transparent objects.
Finally, it is necessary to start integrating object recognition with high-level action planning and memory management so abstract constructs like spatial and temporal relationships can be also recognized.

%
%
%
\subsection{Manipulation}
\label{sec:challenges-man}
Manipulation is perhaps the oldest and most mature research area in robotics, especially when it comes to industrial robots.
With this background, it would be reasonable to expect a similar degree of precision and speed from service robots.
However, mounting a manipulator in a mobile base and limiting available power and computational resources drastically increases the complexity of even the simplest tasks.

\subsubsection*{People's Expectations and Requirements}
In our polls people are explicitly asking for robots able to
clean the toilet, wipe windows, do the dishes (by hand), wash, iron, and fold clothes, open flasks and jars, brush and wash the dog, and take out the garbage
to name some examples.

Also of importance, but not even considered by potential customers are
abilities to open doors, move furniture, operate switches, and to operate the control panel of all electric and electronic appliances.

From these mandatory skills, opening doors has been addressed since 2006 as an optional challenge, but to the date has not been made compulsory in any test.
Nonetheless, door opening was impressively solved by team eR@sers in 2016 with the proprietary robot now used in the \dspl[x].
Notwithstanding, it seems there is still a long way to go before this skill can be considered solved in \opl and \sspl.

\subsubsection*{Advances and Current State}
\begin{figure}[t]
	\centering
	\begin{subfigure}[t]{0.32\columnwidth}
		\centering
		\includegraphics[width=\textwidth,height=3.5cm,keepaspectratio]{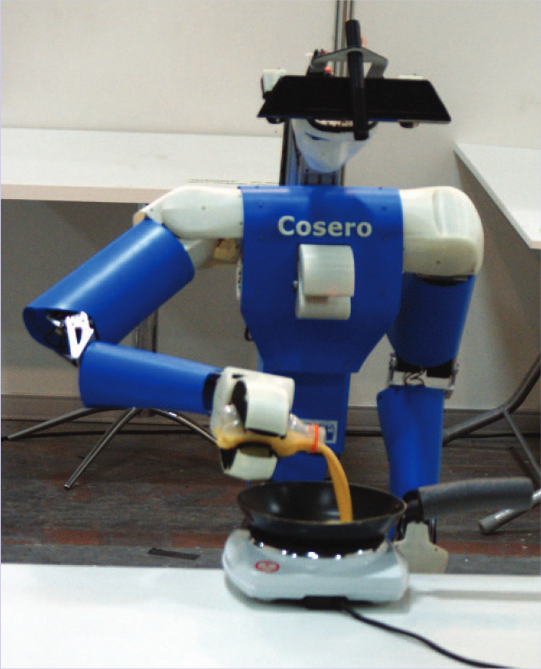}
		\caption{}
		\label{fig:challenge-man-feats-pour}
	\end{subfigure}
	~
	\begin{subfigure}[t]{0.58\columnwidth}
		\centering
		\includegraphics[width=\textwidth,height=3.5cm,keepaspectratio]{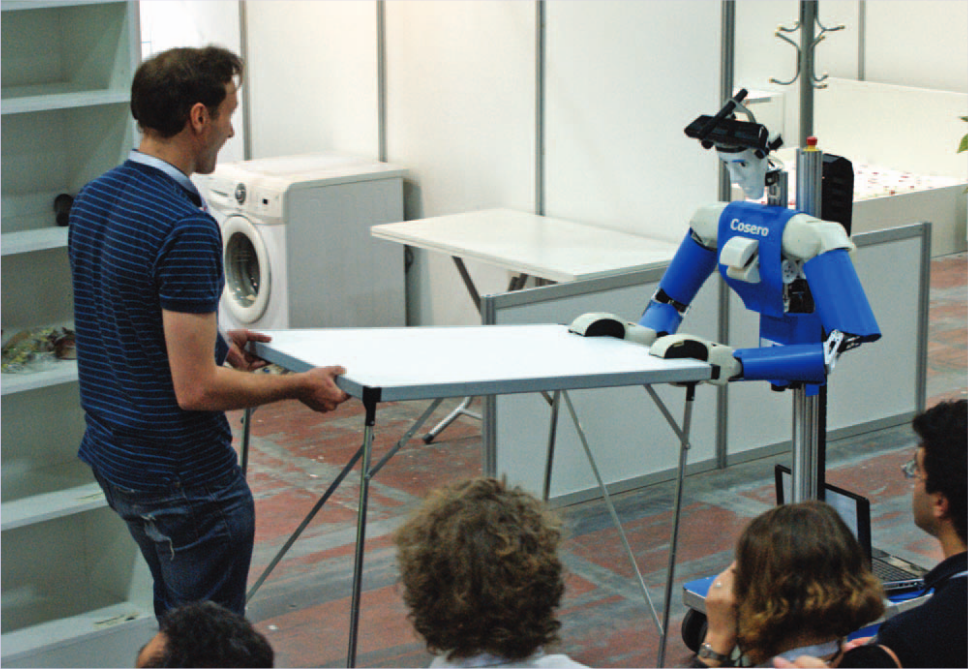}
		\caption{}
		\label{fig:challenge-man-feats-table}
	\end{subfigure}
	\caption{Team Nimbro's final demo in 2011~\Citep{Stuckler2012}. Robot Cosero (a) pouring pancake mixture into a pan, and (b) helping to move a table.}
	\label{fig:challenge-man-feats}
\end{figure}

Unfortunately, the advances in manipulation seem to have stalled~\Citep{Matamoros2018}.
Although challenges like pouring, stacking, transporting a tray, and grasping small objects have been introduced, there have been no serious attempt to solve any of these during regular tests since 2014.
What is worse, in 2017 during the storing groceries test, practically no team attempted to move any object.
Furthermore, none of the aforementioned challenges are new in \athome.
In fact, most of them were already demonstrated several times by the three-times-champion team Nimbro between 2010 and 2014 (see \Cref{fig:challenge-man-feats}), and repeated later on by other teams during final demonstrations.

In addition and compliant with \Cref{sec:challenges-or}, the competition considers mostly moving regular-shaped, small, non-fragile lightweight objects.
Apples, small cereal boxes, and water bottles are some examples of typical objects a robot can be requested to move.
In numbers, most of the objects' weight ranges between 75 and 950 grams.
Dimensions are between 5 and 25 centimeters, with a typical maximum volume of 2 liters.

Also in Stage~I, all robots in the \dspl and the \opl are required to open the door of a cupboard.
This optional task was skipped by all \opl participants in 2017.
More complex tests in Stage~II involve pouring, scrub spots from a table, handling a tray, and moving tableware (bowls, dishes, and cutlery).
Once more, these tasks are often skipped or bypassed, as stated in~\Cref{sec:strategies-man}.

\subsubsection*{Next Steps}
The manipulation capabilities of the robots must be expanded in several directions, including reach, maneuverability, strength, and precision.
Below we describe each of this directions in more detail.

\textbf{Reach:} As of today, robots are required to handle objects in heights ranging from $30cm$ to $1.8m$, but eventually they should be able to reach the floor.
	More important is depth, since today robots are only required to grasp objects at most $5cm$ from the table border.
	Therefore, this distance should be gradually increased until robots can reach objects within an average human grasping distance.

\textbf{Maneuverability:} Nowadays robot's movement is quite clumsy.
	Picking objects from a bag or box and stacking them all together in a narrow space is a trivial task for most people, but difficult even for industrial robots.
	At the same time, twisting, uncapping, shaking, folding, levering, and turning are features yet to be tested in \athome.

\textbf{Strength:} In order to assist most people, a robot should be capable of carrying a 20 liters water bottle, hold a big dog by the leash, and uncap a flask of mayonnaise.
	Moreover, some applications require enough strength to gently lift a fallen elder.
	In consequence, weights and loads should start to gradually increase as soon as possible.

\textbf{Precision:} Today robots are very imprecise.
	To effectively operate in human environments, precision in movement and applied strength is required.
	Like a human, a robot should be able to store a needle and carefully take an egg without crushing it, but also being able to crack it in the right place.
	The same applies to acceleration and speed, either for applying an insulin injection or whipping cream.

Summarizing, some tasks that must be addressed soon within the reach of currently used manipulators include
manipulating a switch,
taking out the bag from the trash bin,
using a ladle to serve soup,
mopping the floor,
dialing a phone number,
picking groceries out of a box,
grasping a towel from a hanger,
and
arranging cutlery into the drawer\footnotemark.

\footnotetext{Several of these tasks have already been proposed for \tc discussion for the next rulebook.}

\subsubsection*{Conclusion}
To foster development in manipulation, several tests were modified for 2018.
Nonetheless, there is still a long way to go before robots can effectively perform in domestic environments.
Obviously, the \athome league would enormously benefit from a manipulation road map with which teams could plan hardware improvements with sufficient time.
For now, it seems sufficient to start by gradually incrementing the weight of most objects, locate them farther away from the corner, and request the robot to perform useful tasks using tools.

%
%
%
\subsection{\asr[x] and \nlp[x]}
\label{sec:challenges-asr}
Speech is a key aspect to people's communication, hence it is fundamental in \hri.
Besides, it is widely acknowledged that language proficiency is related to intelligence.
Therefore, it is expected that robots can understand people's orders expressed naturally, in contrast to the current state where people learn how to talk to a robot.
Regarding spoken interaction, we have identified three key aspects that require attention and are introduced below.

\subsubsection*{Noise}
Noise is one of the most problematic aspects in spoken communication and, in fact, also in competitions like \athome.
The ambient noise produced by several hundred people greatly exceeds the noise levels of an average apartment.
However, so far filtering is delegated to the \asr engine.
Furthermore, eventually robots will be operating in airports, shopping malls, and other noisy environments.
Thus, we think it is best to deal with this issue in an early stage.

\subsubsection*{Non-Native Biased Operators}
Operators in \athome have two fundamental characteristics:
they are robot-friendly and they all have a strong background in robotics.
This means that, unconsciously, operators in \athome are trying to help the robot to succeed.
In other words, they are biased.
The use of biased operators helps robots to perform better and to give a better impression to the audience.
However, it has the disadvantage of voiding the purpose of the competition of providing real-case scenarios for testing.
Furthermore, few operators are native English speakers, which reduces the chances of a robot to deal with the richness of an unconstrained speech production.

\subsubsection*{Generators}
Despite being a powerful tool during the competitions, the use of generators greatly limits the interactions a robot can experience.
In drastical terms, reading verbatim sentences produced by command generators delves \hri into some sort of computer-computer interaction.

\subsubsection*{Next Steps}
A solution to the aforementioned problems has already been proposed.
The next steps consider the use of unbiased people to instruct the robot.
Generators would still be used, but only to select the task, having the operator to command the robot using their own words.
Besides, to address fairness, the interactions should be recorded beforehand, leaving robots to \textit{listen} the recording, but always having the operator at hand for further interactions.
This way, research would be pushed towards free-speech \asr engines, while at the same time, would enforce the integration of the latest techniques in \nlp.
But not only that, this process would lead to the creation of big annotated corpora of spoken \hri for training.

Nonetheless, carefully planned restrictions shall be imposed to the interactions in the first year.
For instance, while in the first years only imperative sentences are used, later on these can be replaced by declarative sentences that depict the desired goal.
Further on, procedures are explained to the robot, reducing the amount of detail with every passing year until the interaction reaches the level of a fluent dialog.

Last, but not least, noise can be introduced gradually by superposing recordings of noisy environments to the ones containing the commands.
This way, all robots would experiment the same initial conditions for the interaction, easing the benchmarking of components with higher levels of abstraction.

%
%
%
\subsection{Roadmaps and milestones}
\label{sec:challenges-roadmaps}
We believe the \athome league has the responsibility of carefully planning what needs to be tested and when to introduce changes by establishing milestones for the competition.
Said in other works, the league should have a set of roadmaps and milestones to asses the \tc when adapting and designing tests.

This is not an unfounded belief.
In fact, this need arises from the teams and the members of the \tc[x] themselves.
To the former, knowing in advance the challenges that will come in future competitions allows to prepare and direct their research.
Moreover, some teams have even stated that they could have made a better use of their resources, should they have known in advance the direction of the league.

With regard to the latter, the \tc renews every year with candidates elected by the community.
In consequence, sometimes it happens that most people are replaced and the direction taken by former members is overridden.
This situation worsens when most of the new members are relatively new in the league (e.g.~less than three years), thus lacking of important undocumented empirical experience.

This lack of direction has led in several occasions to brainstorms in which valuable ideas appear.
These ideas require to be analyzed, evaluated against the robot's capabilities, and condensed in milestones and test drafts that can be retaken in latter years.
Unfortunately, this has not occured until now.

In response, we have condensed not only ideas taken from the teams, but also the people's needs and, considering the current capabilities and limitations of the robots, present them as future steps or milestones in former sections.
Nonetheless, this is just a first step, since roadmaps need to be still prepared and adopted by the league which, in the end, is peer-maintained.

%
%
%
\section{Conclusions and Future Work}
\label{sec:conclusions}
In this paper we conduct a thorough summary of the software solutions and strategies used by participating teams in \athome to address the most important abilities required in the tasks of the competition's tests.
Further on, we present an overview of the attained achievements since the league's foundation based on our experience as long-time participants, contributors, and referees in the league.
Finally, also organized per ability and along with the overview, we discuss these achievements while addressing what is expected by potential consumers, what needs to be done, and would be the next logical steps based on the robot's current capabilities.

This study result in two contributions.
First, we believe the presented summary can serve as quick reference guide for new competitors, or for experienced ones looking for alternatives to their current solutions.
Second, our work sets the basis to build road maps that can help the \athome league, as well as other competitions aiming at service robotics, to plan the direction of the competition towards its goal.
Moreover, we believe that road maps are very important and that planning features to be tested can help teams to prepare in advance, set mid and log-term goals, and have a smarter resource management.
In fact, based on this and previous work, we have designed a roadmap for spoken \hri[x] in \athome which will be presented to the league in Montreal 2018 and in the RoboCup Symposium. 

However, there is still work to be done.
Not only the roadmaps have to be designed.
This work has also allowed us to identify several important flaws that need to be addressed.
For instance, the presence of certain rules might be undermining the development of certain features.
At the same time, we have found that many successful approaches and strategies are never reported in the \tdps.
This has two important setbacks.
First, it makes it much harder for the scientific community to compare the performance of the different approaches when the best performers are missing.
Second, it leads to an eventual loss of knowledge that worsens as the lifetime of a good team shortens.

These insights are left to the competition organizers to analyze as part of future work, for which we trust this manuscript can come handy.

%
%

\bibliographystyle{bib/abbrvnat}
\bibliography{%
	bib/paper,%
	bib/tdps,%
	bib/rulebooks%
}

\begin{thebibliography}{43}
\providecommand{\natexlab}[1]{#1}
\providecommand{\url}[1]{\texttt{#1}}
\expandafter\ifx\csname urlstyle\endcsname\relax
  \providecommand{\doi}[1]{doi: #1}\else
  \providecommand{\doi}{doi: \begingroup \urlstyle{rm}\Url}\fi

\bibitem[Bastianelli et~al.(2014)Bastianelli, Iocchi, Nardi, Castellucci,
  Croce, and Basili]{Bastianelli2014}
E.~Bastianelli, L.~Iocchi, D.~Nardi, G.~Castellucci, D.~Croce, and R.~Basili.
\newblock Robocup@home spoken corpus: using robotic competitions for gathering
  datasets.
\newblock In \emph{Robot Soccer World Cup}, pages 19--30. Springer, 2014.

\bibitem[Bastianelli et~al.(2016)Bastianelli, Croce, Vanzo, Basili, and
  Nardi]{Bastianelli2016}
E.~Bastianelli, D.~Croce, A.~Vanzo, R.~Basili, and D.~Nardi.
\newblock A discriminative approach to grounded spoken language understanding
  in interactive robotics.
\newblock In \emph{IJCAI}, pages 2747--2753, 2016.

\bibitem[Cheng et~al.(2017)Cheng, Ramirez-Amaro, Dianov, Lanillos, Guadarrama,
  Dean, Lozinska, Diez-Valencia, Grzywok, Guo, Simonic, and Wang]{alle2017}
G.~Cheng, K.~Ramirez-Amaro, I.~Dianov, P.~Lanillos, R.~Guadarrama, E.~Dean,
  E.~Lozinska, G.~Diez-Valencia, P.~Grzywok, Q.~Guo, G.~Simonic, and X.~Wang.
\newblock Alle@home 2017 team description paper.
\newblock \emph{RoboCup @Home 2017 Team Description Papers}, 2017.

\bibitem[Christensen et~al.(2017)Christensen, Riek, White, Parashar, Wang,
  Iqbal, Taylor, and Chan]{tritonsdspl2017}
H.~I. Christensen, L.~D. Riek, R.~White, P.~Parashar, S.~Wang, T.~Iqbal,
  A.~Taylor, and D.~Chan.
\newblock Uc san diego 2017 team description paper.
\newblock \emph{RoboCup @Home 2017 Team Description Papers}, 2017.

\bibitem[Demura et~al.(2017)Demura, Demura, Nagashima, Enomoto, Yamakawa,
  Iwasaki, and Mashimo]{happymini2017}
K.~Demura, K.~Demura, K.~Nagashima, K.~Enomoto, T.~Yamakawa, R.~Iwasaki, and
  S.~Mashimo.
\newblock Happy mini 2017 team description paper.
\newblock \emph{RoboCup @Home 2017 Team Description Papers}, 2017.

\bibitem[Dominey(2003)]{Dominey2003}
P.~F. Dominey.
\newblock Learning grammatical constructions from narrated video events for
  human--robot interaction.
\newblock In \emph{Proceedings IEEE humanoid robotics conference, Karlsruhe,
  Germany}, 2003.

\bibitem[Doostdar et~al.(2009)Doostdar, Schiffer, and Lakemeyer]{Doostdar2008}
M.~Doostdar, S.~Schiffer, and G.~Lakemeyer.
\newblock A robust speech recognition system for service-robotics applications.
\newblock In L.~Iocchi, H.~Matsubara, A.~Weitzenfeld, and C.~Zhou, editors,
  \emph{RoboCup 2008: Robot Soccer World Cup XII}, pages 1--12, Berlin,
  Heidelberg, 2009. Springer Berlin Heidelberg.
\newblock ISBN 978-3-642-02921-9.

\bibitem[et~al.(2017)]{walkingmachine2017}
J.~C. et~al.
\newblock Walking machine @home 2017 team description paper.
\newblock \emph{RoboCup @Home 2017 Team Description Papers}, 2017.

\bibitem[Gerlach et~al.(2017)Gerlach, Oschwald, R{\"a}tsch, Ostertag, Weber,
  Huber, Hoffarth, H{\"u}lsmann, Katzenmaier, Kopp, Braun, Foo, Engling, Litz,
  Roericht, Huschitt, Gedrat, Keinert, and Gopal]{rtlions2017}
T.~Gerlach, T.~Oschwald, M.~R{\"a}tsch, F.~Ostertag, T.~Weber, P.~Huber,
  J.~Hoffarth, R.~H{\"u}lsmann, M.~Katzenmaier, P.~Kopp, S.~Braun, M.~Foo,
  J.~Engling, M.~Litz, S.~Roericht, T.~Huschitt, C.~Gedrat, R.~Keinert, and
  A.~Gopal.
\newblock Rt lions team description paper.
\newblock \emph{RoboCup @Home 2017 Team Description Papers}, 2017.

\bibitem[Guo et~al.(2017)Guo, Yao, Ma, Dong, Zhu, Peng, Wang, and
  Ma]{tinker2017}
J.~Guo, H.~Yao, H.~Ma, Y.~Dong, Y.~Zhu, J.~Peng, X.~Wang, and X.~Ma.
\newblock Tinker@home 2017 team description paper.
\newblock \emph{RoboCup @Home 2017 Team Description Papers}, 2017.

\bibitem[Hans et~al.(2008)Hans, Gr{\"a}ser, Jarmer, Schmitt, Bornemeier,
  Lambrecht, and Gossow]{Hans2008}
W.~Hans, S.~Gr{\"a}ser, F.~Jarmer, S.~Schmitt, J.~Bornemeier, P.~Lambrecht, and
  D.~Gossow.
\newblock Robocup 2008 - homer@unikoblenz (germany), 2008.

\bibitem[Hart et~al.(2017)Hart, Stone, Thomaz, and Niekum]{utaustinvilla2017}
J.~W. Hart, P.~Stone, A.~Thomaz, and S.~Niekum.
\newblock Ut austin villa robocup@home domestic standard platform league team
  description paper.
\newblock \emph{RoboCup @Home 2017 Team Description Papers}, 2017.

\bibitem[Hori et~al.(2017{\natexlab{a}})Hori, Ishida, Kiyama, Tanaka, Kuroda,
  Hisano, Imamura, Himaki, Yoshimoto, Aratani, Hashimoto, Iwamoto, Morie, and
  Tamukoh]{hibikinoOPL2017}
S.~Hori, Y.~Ishida, Y.~Kiyama, Y.~Tanaka, Y.~Kuroda, M.~Hisano, Y.~Imamura,
  T.~Himaki, Y.~Yoshimoto, Y.~Aratani, K.~Hashimoto, G.~Iwamoto, T.~Morie, and
  H.~Tamukoh.
\newblock Hibikino-musashi@home 2017 team description paper.
\newblock \emph{RoboCup @Home 2017 Team Description Papers},
  2017{\natexlab{a}}.

\bibitem[Hori et~al.(2017{\natexlab{b}})Hori, Ishida, Kiyama, Tanaka, Kuroda,
  Hisano, Imamura, Himaki, Yoshimoto, Aratani, Hashimoto, Iwamoto, Morie, and
  Tamukoh]{hibikinodspl2017}
S.~Hori, Y.~Ishida, Y.~Kiyama, Y.~Tanaka, Y.~Kuroda, M.~Hisano, Y.~Imamura,
  T.~Himaki, Y.~Yoshimoto, Y.~Aratani, K.~Hashimoto, G.~Iwamoto, T.~Morie, and
  H.~Tamukoh.
\newblock Hibikino-musashi@home spl 2017 team description paper.
\newblock \emph{RoboCup @Home 2017 Team Description Papers},
  2017{\natexlab{b}}.

\bibitem[Iocchi et~al.(2015)Iocchi, Holz, Ruiz-del Solar, Sugiura, and Van
  Der~Zant]{Iocchi2015}
L.~Iocchi, D.~Holz, J.~Ruiz-del Solar, K.~Sugiura, and T.~Van Der~Zant.
\newblock Robocup@home: Analysis and results of evolving competitions for
  domestic and service robots.
\newblock \emph{Artificial Intelligence}, 229:\penalty0 258--281, 2015.

\bibitem[Isobe et~al.(2017)Isobe, Katsumata, Tabuchi, Kinose, Ishimine, Fukui,
  Kobayashi, Taniguchi, Aly, Hagiwara, and Taniguchi]{duckers2017}
S.~Isobe, Y.~Katsumata, Y.~Tabuchi, A.~Kinose, T.~Ishimine, T.~Fukui,
  H.~Kobayashi, A.~Taniguchi, A.~Aly, Y.~Hagiwara, and T.~Taniguchi.
\newblock Duckers 2017 @home dspl team description paper.
\newblock \emph{RoboCup @Home 2017 Team Description Papers}, 2017.

\bibitem[Kelestemur et~al.(2017)Kelestemur, Allaban, and
  Padir]{northeastern2017}
T.~Kelestemur, A.~A. Allaban, and T.~Padir.
\newblock Frasier: Fostering resilient aging with self-efficacy and
  independence enabling robot team northeastern’s approach for robocup@home.
\newblock \emph{RoboCup @Home 2017 Team Description Papers}, 2017.

\bibitem[Lee et~al.(2017)Lee, Choi, Lee, Park, Choi, Baek, and
  Zhang]{aupair2017}
B.-J. Lee, J.-Y. Choi, C.-Y. Lee, K.-W. Park, S.~Choi, C.~Baek, and B.-T.
  Zhang.
\newblock 2017 aupair team description paper.
\newblock \emph{RoboCup @Home 2017 Team Description Papers}, 2017.

\bibitem[Li et~al.(2017)Li, Liu, Su, Li, Cheng, Hsieh, Liang, Chen, Lin, and
  Chang]{airobots2017}
T.-H.~S. Li, C.-Y. Liu, Y.-T. Su, C.-H. Li, C.-W. Cheng, C.-Y. Hsieh, J.-J.
  Liang, C.-Y. Chen, H.-Y. Lin, and K.-C. Chang.
\newblock airobots\_ncku 2017 team description paper.
\newblock \emph{RoboCup @Home 2017 Team Description Papers}, 2017.

\bibitem[Liu et~al.(2017)Liu, Zhang, Tang, and Chen]{wrighteagle2017}
J.~Liu, Z.~Zhang, B.~Tang, and X.~Chen.
\newblock Wrighteagle@home 2017 team description paper.
\newblock \emph{RoboCup @Home 2017 Team Description Papers}, 2017.

\bibitem[Mart{\i}nez et~al.(2015)Mart{\i}nez, Pavez, Olave, Correa,
  S{\'a}nchez, Loncomilla, and Ruiz-del Solar]{uchile2015}
L.~Mart{\i}nez, M.~Pavez, G.~Olave, M.~Correa, L.~S{\'a}nchez, P.~Loncomilla,
  and J.~Ruiz-del Solar.
\newblock Uchile homebreakers 2015 team description paper.
\newblock 2015.

\bibitem[Mart{\'i}nez et~al.(2017)Mart{\'i}nez, Mu{\~n}oz, Olave, Pais, Hernan,
  Gomez, Garrido, Campanini, Orellana, Loncomilla, and Ruiz-del
  Solar]{uchile2017}
L.~Mart{\'i}nez, R.~Mu{\~n}oz, G.~Olave, G.~Pais, G.~Hernan, D.~Gomez,
  L.~Garrido, D.~Campanini, P.~Orellana, P.~Loncomilla, and J.~Ruiz-del Solar.
\newblock Uchile homebreakers 2017 team description paper.
\newblock \emph{RoboCup @Home 2017 Team Description Papers}, 2017.

\bibitem[Matamoros et~al.(2018{\natexlab{a}})Matamoros, Harbusch, and
  Paulus]{Matamoros2018rc}
M.~Matamoros, K.~Harbusch, and D.~Paulus.
\newblock From commands to goal-based dialogs: A roadmap to achieve natural
  language interaction in robocup@home.
\newblock 2018{\natexlab{a}}.

\bibitem[Matamoros et~al.(2018{\natexlab{b}})Matamoros, Seib, Memmesheimer, and
  Paulus]{Matamoros2018}
M.~Matamoros, V.~Seib, R.~Memmesheimer, and D.~Paulus.
\newblock Robocup@home: Summarizing achievements in over eleven years of
  competition.
\newblock In \emph{2018 IEEE International Conference on Autonomous Robot
  Systems and Competitions (ICARSC)}, pages 186--191, April 2018{\natexlab{b}}.
\newblock \doi{10.1109/ICARSC.2018.8374181}.

\bibitem[Memmesheimer et~al.(2017)Memmesheimer, Wettengel, Daniel~Müller,
  Roosen, Buchhold, Moritz~Löhne, Mykhalchyshyna, and Paulus]{homer2017}
R.~Memmesheimer, N.~Y. Wettengel, F.~Daniel~Müller, Polster, M.~Roosen,
  L.~Buchhold, M.~Moritz~Löhne, Schnorr, I.~Mykhalchyshyna, and D.~Paulus.
\newblock Robocup 2017 - homer@unikoblenz (germany).
\newblock \emph{RoboCup @Home 2017 Team Description Papers}, 2017.

\bibitem[Miyawaki et~al.(2017)Miyawaki, Sano, Inoue, Hiroi, Nishiguchi, and
  Suzuki]{oittrial2017}
K.~Miyawaki, M.~Sano, Y.~Inoue, Y.~Hiroi, S.~Nishiguchi, and M.~Suzuki.
\newblock O.i.t-trial 2017 team description paper.
\newblock \emph{RoboCup @Home 2017 Team Description Papers}, 2017.

\bibitem[M.T.~Lázaro et~al.(2017)M.T.~Lázaro, Nardi, Hanheide, and
  Fentanes]{spqrel2017}
L.~M.T.~Lázaro, Iocchi, D.~Nardi, M.~Hanheide, and J.~P. Fentanes.
\newblock Spqrel 2017 team description paper.
\newblock \emph{RoboCup @Home 2017 Team Description Papers}, 2017.

\bibitem[Oishi et~al.(2017)Oishi, Miura, Koide, Demura, Kohari, Une,
  Villamar-Gomez, Kato, Kojima, and Morohashi]{aisltut2017}
S.~Oishi, J.~Miura, K.~Koide, M.~Demura, Y.~Kohari, S.~Une, L.~Villamar-Gomez,
  T.~Kato, M.~Kojima, and K.~Morohashi.
\newblock Aisl-tut @home league 2017 team description paper.
\newblock \emph{RoboCup @Home 2017 Team Description Papers}, 2017.

\bibitem[Okada et~al.(2017)Okada, Yokoyama, Ogata, Inamura, Iwahashi, and
  Sugiura]{erasers2017}
H.~Okada, H.~Yokoyama, T.~Ogata, T.~Inamura, N.~Iwahashi, and K.~Sugiura.
\newblock Team er@sers[dspl] (toyota hsr) 2017 team description paper.
\newblock \emph{RoboCup @Home 2017 Team Description Papers}, 2017.

\bibitem[Oliva and Torralba(2007)]{Oliva2007}
A.~Oliva and A.~Torralba.
\newblock The role of context in object recognition.
\newblock \emph{Trends in cognitive sciences}, 11\penalty0 (12):\penalty0
  520--527, 2007.

\bibitem[Oliva et~al.(2004)Oliva, Wolfe, and Arsenio]{Oliva2004}
A.~Oliva, J.~M. Wolfe, and H.~C. Arsenio.
\newblock Panoramic search: the interaction of memory and vision in search
  through a familiar scene.
\newblock \emph{Journal of Experimental Psychology: Human Perception and
  Performance}, 30\penalty0 (6):\penalty0 1132, 2004.

\bibitem[Pineda et~al.(2017)Pineda, Rasc{\'o}n, Fuentes, Rodr{\'i}guez, Ortega,
  Reyes, Hern{\'a}ndez, Cruz, V{\'e}lez, and Ram{\'i}rez]{golem2017}
L.~A. Pineda, C.~Rasc{\'o}n, G.~Fuentes, A.~Rodr{\'i}guez, H.~Ortega, M.~Reyes,
  N.~Hern{\'a}ndez, R.~Cruz, I.~V{\'e}lez, and M.~Ram{\'i}rez.
\newblock The golem team, robocup@home 2017.
\newblock \emph{RoboCup @Home 2017 Team Description Papers}, 2017.

\bibitem[Roy and Pentland(2002)]{Roy2002}
D.~K. Roy and A.~P. Pentland.
\newblock Learning words from sights and sounds: A computational model.
\newblock \emph{Cognitive science}, 26\penalty0 (1):\penalty0 113--146, 2002.

\bibitem[Sammut et~al.(2017)Sammut, Pagnucco, Castro, Flannagan, Gratton,
  Hengst, Rajaratnam, Schwering, Thielscher, Velonaki, and Wiley]{unsw2017}
C.~Sammut, M.~Pagnucco, G.~Castro, C.~Flannagan, M.~Gratton, B.~Hengst,
  D.~Rajaratnam, C.~Schwering, M.~Thielscher, M.~Velonaki, and T.~Wiley.
\newblock Unsw robocup@home spl team description paper.
\newblock \emph{RoboCup @Home 2017 Team Description Papers}, 2017.

\bibitem[Savage et~al.(2017)Savage, Negrete, Cruz, Marquez, Martell, Cruz,
  Vazquez, Pano, Cruz, Silva, Estrada, Arce, Matamoros, Garzon, and
  Fuentes]{pumas2017}
J.~Savage, M.~Negrete, J.~Cruz, J.~Marquez, R.~Martell, J.~Cruz, E.~Vazquez,
  M.~Pano, J.~Cruz, E.~Silva, H.~Estrada, H.~Arce, M.~Matamoros, A.~Garzon, and
  O.~Fuentes.
\newblock Pumas@home 2017 team description paper.
\newblock \emph{RoboCup @Home 2017 Team Description Papers}, 2017.

\bibitem[Seib et~al.(2015)Seib, Manthe, Memmesheimer, Polster, and
  Paulus]{Seib2015}
V.~Seib, S.~Manthe, R.~Memmesheimer, F.~Polster, and D.~Paulus.
\newblock Team homer@unikoblenz—approaches and contributions to the
  robocup@home competition.
\newblock In \emph{Robot Soccer World Cup}, pages 83--94. Springer, 2015.

\bibitem[Stanford(2011)]{stanford2011corenlponline}
Stanford.
\newblock Corenlp, 2011.
\newblock URL \url{http://nlp.stanford.edu:8080/corenlp/}.

\bibitem[Stuckler et~al.(2012)Stuckler, Holz, and Behnke]{Stuckler2012}
J.~Stuckler, D.~Holz, and S.~Behnke.
\newblock Robocup@home: Demonstrating everyday manipulation skills in robocup
  @home.
\newblock \emph{IEEE Robotics \& Automation Magazine}, 19\penalty0
  (2):\penalty0 34--42, 2012.

\bibitem[Tan et~al.(2017)Tan, Duan, Ismail, and Uchimura]{kamerideropl2017}
J.~T.~C. Tan, F.~Duan, Z.~H.~b. Ismail, and Y.~Uchimura.
\newblock Kamerider opl @home 2017 team description paper.
\newblock \emph{RoboCup @Home 2017 Team Description Papers}, 2017.

\bibitem[van~der Burgh et~al.(2017)van~der Burgh, Lunenburg, Appeldoorn,
  Wijnands, Clephas, Baeten, van Beek, Ottervanger, van Rooy, and van~de
  Molengraft]{tue2017}
M.~van~der Burgh, J.~Lunenburg, R.~Appeldoorn, R.~Wijnands, T.~Clephas,
  M.~Baeten, L.~van Beek, R.~Ottervanger, H.~van Rooy, and M.~van~de
  Molengraft.
\newblock Tech united eindhoven @home 2017 team description paper.
\newblock \emph{University of Technology Eindhoven}, 2017.

\bibitem[Wachsmuth et~al.(2017)Wachsmuth, Lier, Meyer~zu Borgsen, Kummert,
  Lach, and Sixt]{tobi2017}
S.~Wachsmuth, F.~Lier, S.~Meyer~zu Borgsen, J.~Kummert, L.~Lach, and D.~Sixt.
\newblock Tobi-team of bielefeld: The human-robot interaction system for
  robocup @home 2017.
\newblock 2017.

\bibitem[Wisspeintner et~al.(2009)Wisspeintner, Van Der~Zant, Iocchi, and
  Schiffer]{Wisspeintner2009}
T.~Wisspeintner, T.~Van Der~Zant, L.~Iocchi, and S.~Schiffer.
\newblock Robocup@home: Scientific competition and benchmarking for domestic
  service robots.
\newblock \emph{Interaction Studies}, 10\penalty0 (3):\penalty0 392--426, 2009.

\bibitem[Yaguchi et~al.(2017)Yaguchi, Tran, Takeda, Kochigami, Li, Sasabuchi,
  Furuta, Nagahama, Okada, and Inaba]{jsk2017}
H.~Yaguchi, B.~Tran, M.~Takeda, K.~Kochigami, Z.~Li, K.~Sasabuchi, Y.~Furuta,
  K.~Nagahama, K.~Okada, and M.~Inaba.
\newblock Jsk@home: Team description paper for robocup@home 2017.
\newblock \emph{RoboCup @Home 2017 Team Description Papers}, 2017.

\end{thebibliography}


\end{document}